\pdfoutput=1
\documentclass[10pt,twocolumn,letterpaper]{article}

\usepackage[pagenumbers]{cvpr}

\usepackage{graphicx}
\usepackage{stfloats}
\usepackage{amsmath}
\usepackage{amssymb}
\usepackage{booktabs}
\usepackage{multirow}
\usepackage{array}
\usepackage{tabularx}
\usepackage{xcolor}
\usepackage{colortbl}
\usepackage{url}
\usepackage{float}
\definecolor{cvprblue}{rgb}{0.21,0.49,0.74}
\usepackage[pagebackref,breaklinks,colorlinks,allcolors=cvprblue]{hyperref}

\newcommand{\dataset}{\textit{ForkTrack-Bench}}
\newcommand{\method}{\textit{TrackWAM}}

\title{
TrackWAM: Learning Predictive Dynamics for Embodied Visual Tracking
}

\author{
Zeying Gong$^{1}$ \quad Heting Ying$^{2}$ \quad Peng Liu$^{2}$ \quad Yangyi Zhong$^{1}$ \\
Rong Li$^{1}$ \quad Tiancheng Zhao$^{2,3\,\dagger}$ \quad Junwei Liang$^{1,4\,\dagger}$ \\[3pt]
{\normalsize $^{1}$The Hong Kong University of Science and Technology (Guangzhou) \quad $^{2}$Om AI Research} \\
{\normalsize $^{3}$Binjiang Institute of Zhejiang University \quad $^{4}$The Hong Kong University of Science and Technology} \\[3pt]
{\normalsize Project Page: \url{https://zeying-gong.github.io/TrackWAM-Web/}}
}

\begin{document}

\maketitle
\begingroup
\renewcommand\thefootnote{}
\begin{NoHyper}\footnotetext{$^{\dagger}$Corresponding authors.}\end{NoHyper}
\endgroup

\begin{abstract}
Embodied visual tracking requires a robot to follow a moving
target from egocentric observations.
Visually similar distractors and occlusions create perceptual ambiguity, while uncertainty in target motion complicates planning. Inference latency further limits how quickly the robot can adapt.
We propose TrackWAM, a lightweight tracking framework that
combines spatial supervision with prediction-conditioned control.
Its Track Perceiver Blocks fuse the target template with visual
history to form target-conditioned spatiotemporal representations.
The World Expert combines these representations with current
and predicted visual features to estimate the target's current
and future spatial states.
The Action Expert uses both states and their latent representations to generate a short robot trajectory.
We also introduce ForkTrack-Bench, a benchmark with controlled target--distractor appearance similarity and alternative target trajectories that branch from shared observation--action histories.
TrackWAM achieves state-of-the-art performance on both
EVT-Bench and ForkTrack-Bench, with lower inference latency
than all evaluated open-source MLLM-based baselines.
On the Unitree Go2 robot, TrackWAM achieves an overall success
rate of 80.0\% and an average inference latency of 100.2~ms
across indoor layouts, outdoor walkways, and crowded scenes.
Code and dataset are available on our project page.
\end{abstract}

\section{Introduction}
Embodied visual tracking (EVT) requires an embodied agent to continuously follow a specified target using visual observations~\cite{maalouf2024follow,zhang2018coarse,zhong2019advat,zhong2023rspt,zhong2024offlineevt,trackvla2025}. Compared with tracking in recorded videos~\cite{babenko2009visual,ye2022ostrack,fan2019lasot}, the agent's own actions alter subsequent viewpoints, making the target's appearance, position, and visibility depend on both target behavior and camera motion~\cite{zhong2019advat,zhong2021advatplus,zhong2023rspt,zhong2024offlineevt,trackvla2025,oavat2026}. EVT thus brings together video understanding, target tracking, and action generation.

Two challenges limit embodied visual tracking in dynamic environments. \textbf{1) Perceptual ambiguity.} Similar-looking pedestrians can introduce severe identity ambiguity, while temporary occlusions render historical position estimates stale~\cite{maalouf2024follow,zhong2024offlineevt}. Although recent Multimodal Large Language Model-based (MLLM-based) trackers mitigate this through spatial reasoning and temporal memory~\cite{trackvla2025,trackvlapp2025,refertrack2026}, maintaining reliable tracking in a closed loop still demands robust target-state estimation under continuous ego-motion and viewpoint changes. \textbf{2) Planning uncertainty.} Once the target's trajectory diverges, the policy must rapidly update its actions upon receiving new visual evidence. High latency compounds this challenge by widening the temporal gap between observation and action.

\begin{figure}[t]
    \centering
    \includegraphics[width=\columnwidth]{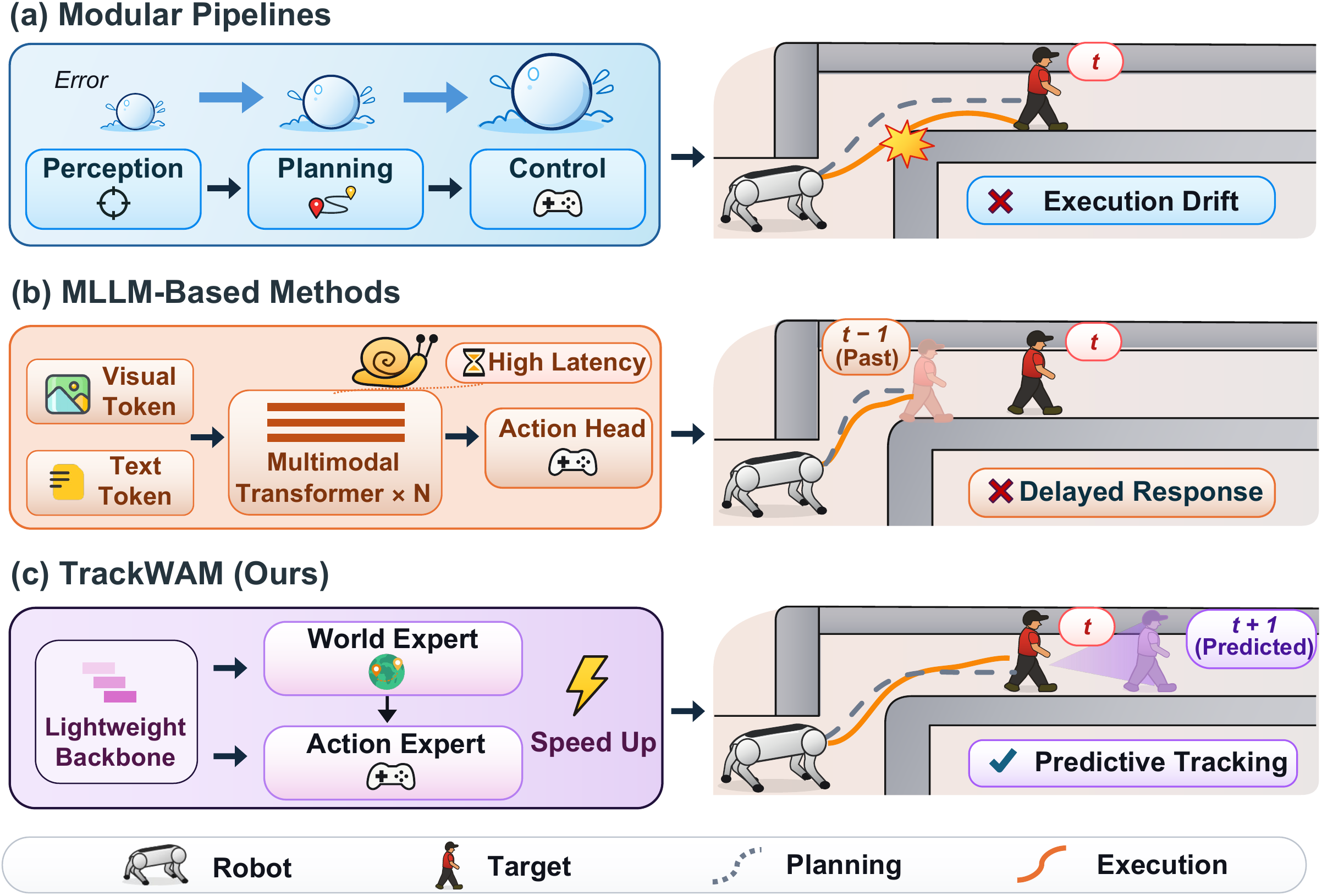}
\caption{\textbf{Motivation of \method{}.}
(a) Perception-planning error drift.
(b) Reaction delays from MLLM latency.
(c) \method{} addresses these via world-conditioned action forecasting.}
    \label{fig:teaser}
    \vspace{-1.5em}
\end{figure}

While existing benchmarks~\cite{qiu2016unrealcv, trackvla2025, adatracker2026} support closed-loop evaluation or cross-embodiment transfer, they are not designed to systematically diagnose the two core challenges of embodied tracking: perceptual ambiguity under appearance-similar distractors and planning uncertainty under divergent target futures. To bridge this gap, we introduce \dataset{}, a novel benchmark designed to diagnose embodied tracking across two critical dimensions. First, to address perceptual ambiguity, \dataset{} quantifies and controls the appearance similarity of distractor pedestrians relative to the target. Second, to address planning uncertainty, it constructs counterfactual branching trajectories where parallel tracking episodes diverge from identical observation--action histories into alternative target futures. Therefore, \dataset{} isolates perceptual confounds and target intent variations.

To address these two challenges, we propose \method{}, a lightweight
embodied tracking framework that couples target-aware perception
with spatial state prediction and closed-loop control.
The \textbf{Track Perceiver Blocks} use learnable queries and bidirectional
attention to fuse the target template with visual history,
forming target-conditioned spatiotemporal representations.
The \textbf{World Expert} combines these representations with current
and predicted visual features to estimate the target's current
and future spatial states under explicit spatial supervision.
The \textbf{Action Expert} then integrates the fused queries with the
current and predicted target representations to jointly generate
a short robot trajectory, enabling closed-loop tracking.

\method{} achieves state-of-the-art (SOTA) performance on both EVT-Bench and \dataset{}. Under identical hardware configurations, it has lower inference latency than all open-source MLLM-based baselines. On a Unitree Go2, \method{} achieves an overall success rate of 80.0\% with an average inference latency of 100.2~ms across indoor layouts, outdoor walkways, and crowded scenes.

Our contributions are as follows:

\begin{itemize}
    \item \textbf{A World-Action Modeling Framework.} We propose \method{}, a lightweight framework that couples predictive perception with motion planning. The shared temporal representation preserves target identity across frames and connects future prediction with embodied control.
    \item \textbf{A Controllable Multi-Future Dataset.} We introduce \dataset{} to support both policy training and evaluation under perceptual ambiguity and planning uncertainty, while separating perceptual confusion from counterfactual target futures.
    \item \textbf{Simulation and Real-World Evaluation.} Extensive experiments across benchmarks and physical deployment on a Unitree Go2 robot demonstrate that \method{} achieves superior tracking accuracy, real-time efficiency, and robustness in complex dynamic environments.
\end{itemize}

\section{Related Work}
\subsection{Vision-Language-Action Models for Tracking}

Recently, Vision-Language-Action (VLA) models have introduced strong zero-shot reasoning into tracking and target-following tasks. TrackVLA~\cite{trackvla2025} and TrackVLA++~\cite{trackvlapp2025} unify visual target recognition with spatial trajectory prediction, while OmTrackVLA~\cite{lee2026vlxgorobustembodiedtracking} predicts short-horizon waypoints from monocular video and language instructions. ReferTrack~\cite{refertrack2026} grounds a language instruction in an indexed set of pedestrian detections, conditions waypoint prediction on the selected box, and retains previously selected boxes as target history. In parallel, general-purpose navigation backbones adapt to target following by integrating multi-modal prompts: Uni-NaVid~\cite{zhang2024uninavid} and NavFoM~\cite{zhang2025navfom} encode continuous video streams to predict action tokens, whereas models like Qwen-RobotNav~\cite{qwenrobotnav2026}, ABot-N0~\cite{abotn02026}, and LightNav-0~\cite{lightnav2026} formulate following as high-level visual instruction navigation. \method{} leverages predicted feature priors to capture temporal-spatial dynamics and facilitate robust action generation, enabling the action head to operate on target-conditioned representations.

\subsection{World-Action Models for Embodied Control}

To complement the lack of spatial-temporal foresight in VLA policies, world models~\cite{ha2018worldmodels} learn environmental dynamics to guide control. In model-based RL, DreamerV3~\cite{hafner2025dreamerv3} learns compact latent dynamics, TD-MPC2~\cite{hansen2024tdmpc2} combines implicit world models with local trajectory optimization, and DayDreamer~\cite{wu2023daydreamer} adapts world-model learning directly to physical robots. For visual navigation, Pathdreamer~\cite{koh2021pathdreamer} synthesizes multi-view panoramic RGB futures, while Navigation World Models~\cite{bar2025nwm} predict video sequences to model scene dynamics. Shifting from frame generation to representation learning, NavMorph~\cite{yao2025navmorph} aligns predicted features via deformable motion fields, NavForesee~\cite{liu2026navforesee} predicts long-term spatial connectivity for target search, and FutureNav~\cite{zhang2026futurenav} uses action-conditioned visual feature extrapolation to guide local path planning. Nevertheless, video generators introduce high computation, and latent planners often require continuous online rollout. Instead, \method{} uses predicted feature priors to jointly model spatio-temporal dynamics for real-time control.

\section{Problem Formulation}
\label{sec:task_definition}

\begin{figure*}[t]
    \centering
    \includegraphics[width=\linewidth]{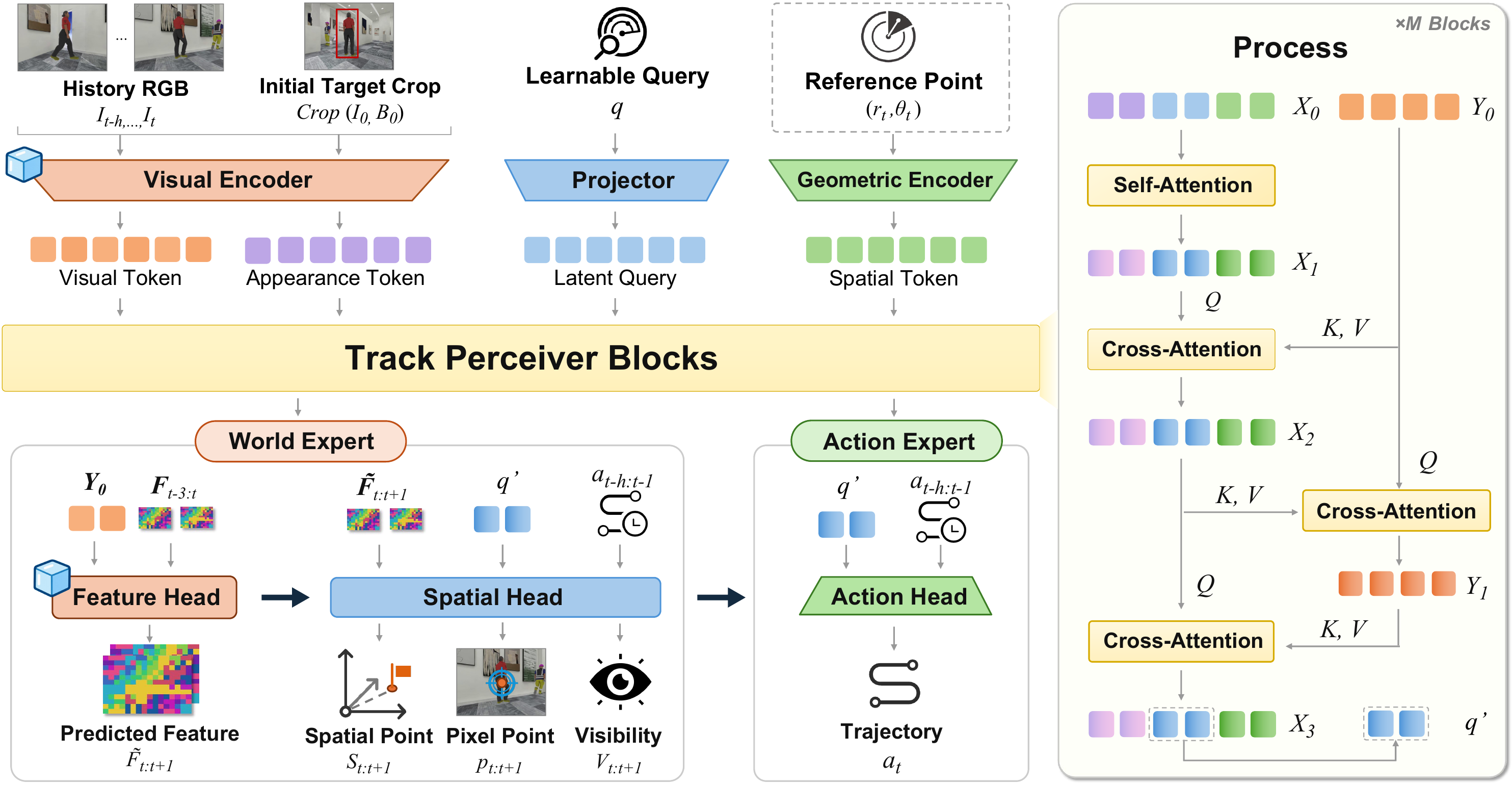}
    \caption{\textbf{Overview of \method{}.} \textbf{(1) Track Perceiver Blocks} combine target cues, visual history, temporal context, and learned queries to estimate the current target state. \textbf{(2) World Expert} forecasts future visual features and target spatial states to provide spatio-temporal foresight. \textbf{(3) Action Expert} conditions on both current and predicted states to generate a sequence of waypoints for predictive tracking.}
    \label{fig:trackwam_overview}
    \vspace{-1.5em}
\end{figure*}

EVT aims to control a mobile robot to follow a
specified target instance in a dynamic 3D environment. We formulate EVT as a
partially observed, closed-loop control problem.

\noindent\textbf{Observations.}
At each step $t$, the robot receives an RGB image $I_t$, with the target specified by an initial bounding box $B_0$ in $I_0$ (see Appendix~\ref{sec:supp_target_init} for initialization details).

\noindent\textbf{Actions.}
At each step $t$, the Action Expert outputs a short trajectory
$a_t$ of planar waypoints in the robot frame. For execution,
the controller uses these waypoints to compute the robot's
forward velocity, lateral velocity, and yaw rate.

\noindent\textbf{EVT policy.}
Let $\mathcal{I}^{(h+1)}_t$ denote a visual history of $h$ past
frames and the current frame $I_t$. Our policy $\pi_\theta$
uses this history, the initial target crop, and the executed
action history to predict a robot-frame trajectory $a_t$.
\begin{equation}
    a_t = \pi_\theta\!\left(
        \mathcal{I}^{(h+1)}_t,\operatorname{Crop}(I_0,B_0),a_{t-h:t-1}
    \right).
    \label{eq:evt_tracker}
\end{equation}

\noindent\textbf{Success criterion.} An episode succeeds if, at its end, the robot remains oriented toward the target and maintains a safe distance $d \in [d_{min}, d_{max}]$ (\textit{e.g.}, $1$–$3$ meters).

\section{Method}
\label{sec:method}

\textbf{Overview.}
We introduce \method{}, as shown in Fig.~\ref{fig:trackwam_overview}. It is a framework with the following modules. \textbf{(1) Track Perceiver Blocks.} To tackle perceptual ambiguity, this module fuses the target template
with visual history and temporal context using learnable queries to form a target-aware representation.
\textbf{(2) World Expert.} To anticipate environment and target dynamics, this module forecasts next-frame visual features and future target spatial states, providing essential spatio-temporal foresight.
\textbf{(3) Action Expert.} To mitigate planning uncertainty, this module conditions on both current and predicted target states to generate actions.

\subsection{Track Perceiver Blocks}
\label{sec:method_perceiver}

\noindent\textbf{Visual memory construction.}
To capture temporal dynamics and spatial detail at time $t$, a shared encoder $\phi$ encodes observations $I_{t-h:t}$ and the initial target crop $B_0$. TrackWAM uses a 32-frame RGB window ($h=31$), including the current frame. A trainable projection maps patch features to a common dimension. To balance efficiency and precision, historical images are pooled onto a coarse grid, whereas the current frame retains a fine grid, and crop features form template tokens $E^{\mathrm{tar}}$.

We extract one global token per observed frame and add a
relative-time encoding and a learned temporal position embedding.
A causal Transformer produces a temporal representation
$\mathbf{e}_i^{\mathrm{time}}$ for each frame $i$, attending
only to observations at or before that frame.
These representations are added back to the corresponding
spatial tokens to form the spatiotemporal visual memory $Y_0$ for the first block.

\noindent\textbf{Target-aware fusion.}
Inspired by Perceiver~\cite{jaegle2021perceiver}, we use learnable queries and target template
tokens to jointly guide visual feature fusion, yielding a
compact, target-conditioned representation of the observed scene
and its temporal context. We stack $M=3$ Track Perceiver Blocks.
Each block takes tracking queries $q$ and visual memory $Y_0$.
In the first block, $q$ consists of $N_q$ learnable queries.
Self-attention fuses the queries with the target template, and
bidirectional cross-attention updates both token streams.
\begin{align}
    X_0 &= [q;E^{\mathrm{tar}}],
    & X_1 &= \operatorname{SA}(X_0), \\
    X_2 &= \operatorname{CA}(X_1,Y_0),
    & Y_1 &= \operatorname{CA}(Y_0,X_2), \\
    X_3 &= \operatorname{CA}(X_2,Y_1),
    & q' &= \operatorname{Select}_{q}(X_3).
\end{align}
Here $[\,\cdot\,;\,\cdot\,]$ denotes token concatenation,
$\operatorname{SA}$ denotes self-attention, and
$\operatorname{CA}(A,B)$ uses $A$ as queries and $B$ as keys
and values. $\operatorname{Select}_{q}$ retains the $N_q$
tracking query positions. Each block passes $q'$ and $Y_1$
to the next block as its $q$ and $Y_0$. We denote the query
output of block $j$ by $q_j$ and mean-pool the final queries
to obtain $\bar{q}$.

In the UWB ablation, a geometric encoder maps the target range
and bearing $u_t=(r_t,\theta_t)$ to spatial tokens, which we
append to $X_0$ before self-attention.

\subsection{World Expert}
\label{sec:method_world}

\noindent\textbf{Current and future visual features.}
To anticipate environmental dynamics and localize the target
over time, the World Expert estimates the current target state
and predicts its future state by combining target-conditioned
representations with current and predicted visual features.
Let $F_i$ denote the visual features extracted from observation
$I_i$ by the DINO encoder $\phi$.
Given the features of the $N_p=4$ most recent observations,
a frozen pretrained predictor $\mathcal{P}$ forecasts the
next-frame feature map:
\begin{equation}
    \widetilde{F}_{t+1}
    = \mathcal{P}\!\left(F_{t-N_p+1:t}\right).
    \label{eq:world_prediction}
\end{equation}
We project the current features $F_t$ and predicted features
$\widetilde{F}_{t+1}$ into the feature dimension used by the
Track Perceiver Blocks, obtaining spatial maps $C_t$ and
$\hat{C}_{t+1}$, respectively.
The predictor uses only observed visual features, without
access to the future image $I_{t+1}$.

\noindent\textbf{Target spatial state readout.}
To identify the designated target within each spatial map,
we condition the state readouts on the fused queries,
target template, and temporal context.
The current-state readout estimates the target state from $C_t$,
while the future-state readout predicts its next-step state
from $\hat{C}_{t+1}$.
The future-state readout additionally incorporates the history
of executed control commands, providing ego-motion context
for future target state prediction:
\begin{align}
    (z_t,\hat{s}_t)
    &= R_{\mathrm{cur}}\!\left(
        \bar{q},\mathbf{e}_t^{\mathrm{time}},
        \bar{E}^{\mathrm{tar}},C_t\right), \\
    (\hat{z}_{t+1},\hat{s}_{t+1})
    &= R_{\mathrm{fut}}\!\left(
        \bar{q},\mathbf{e}_t^{\mathrm{time}}+a_{t-h:t-1},
        \bar{E}^{\mathrm{tar}},\hat{C}_{t+1}\right),
    \label{eq:target_state_readout}
\end{align}
where $\bar{q}$ and $\bar{E}^{\mathrm{tar}}$ denote the pooled
final queries and template features, respectively, and $a_{t-h:t-1}$
encodes previously executed control commands.
$R_{\mathrm{cur}}$ and $R_{\mathrm{fut}}$ are the current-
and future-state readouts, producing latent representations
$z_t$ and $\hat{z}_{t+1}$ alongside their spatial state estimates.

Both readouts predict target visibility $V_{t:t+1}$, an image-plane point distribution $p_{t:t+1}$ (with a dedicated invisible class), and a robot-frame 3D point $S_{t:t+1}$ with position uncertainties. While the visual predictor remains frozen, the readouts are trained jointly with the policy, conditioned on command history for the future state. The current estimate $\hat{s}_t$, future prediction $\hat{s}_{t+1}$, and their latent representations are then provided to the Action Expert.

\subsection{Action Expert}
\label{sec:method_action}

\noindent\textbf{Action conditioning.}
The Action Expert uses the fused queries and the current
and predicted target states from the World Expert, together
with their latent representations, to generate action $a_t$.
We construct the action context
\begin{equation}
\begin{aligned}
g_t^{\mathrm{ctx}}=\bigl[
    &\bar{q};\,\mathbf{e}_t^{\mathrm{time}};\,a_{t-h:t-1};\,
    z_t;\,\hat{z}_{t+1};\,m_t;\\
    &r(\hat{s}_t);\,r(\hat{s}_{t+1});\,
    r(\hat{s}_t)-r(\hat{s}_{t+1})
\bigr],
\end{aligned}
\label{eq:action_context}
\end{equation}
where $\bar{q}$ is the mean of the final queries,
$m_t$ summarizes the pooled current and predicted spatial maps
$C_t$ and $\hat{C}_{t+1}$, and $r(\cdot)$ converts the target-state
predictions and uncertainties into a compact vector.
Learned projections map $g_t^{\mathrm{ctx}}$ to a shared context
token and each query output $q_j$ to stage-specific query tokens.
Together, these form the conditioning tokens $\mathcal{G}_t$
used by successive layers of a time-conditioned flow Transformer.

\noindent\textbf{Trajectory generation.}
The generated sequence contains $N_a=8$ tokens, each representing
a normalized planar pose increment $(\Delta x,\Delta y,\Delta\theta)$
over a 0.1~s interval. The poses and increments are expressed
in the robot frame fixed at time $t$.

We generate the sequence using conditional flow
matching~\cite{lipman2023flow}.
Let $X_{t,\tau}$ denote the trajectory-token sequence at flow time
$\tau\in[0,1]$.
Starting from Gaussian noise, the sequence evolves according to
\begin{equation}
\begin{aligned}
    \frac{\mathrm{d}X_{t,\tau}}{\mathrm{d}\tau}
    &= f_\theta(X_{t,\tau},\tau,\mathcal{G}_t),\\
    X_{t,0} &\sim \mathcal{N}(0,I),
\end{aligned}
\label{eq:trajectory_flow}
\end{equation}
where $f_\theta$ predicts a normalized-trajectory velocity field.

At inference, we denormalize and cumulatively sum the predicted
pose increments to obtain the robot-frame trajectory $a_t$.
The policy replans when the next observation arrives.

\section{\dataset{} Dataset}
\label{sec:dataset}

\begin{figure*}[t]
    \centering
    \includegraphics[width=\linewidth]{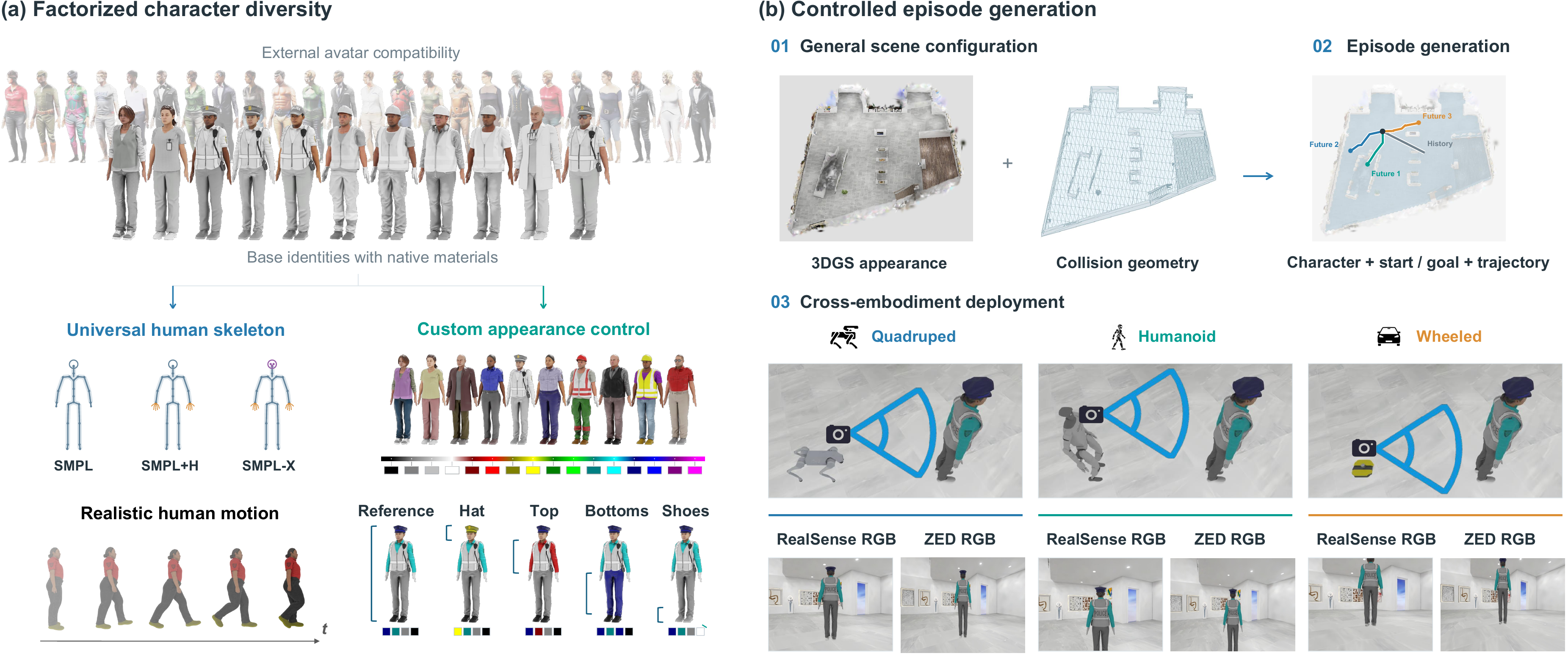}
    \caption{\textbf{Data generation pipeline of \dataset{}.}
(a) The assets support cross-compatibility, universal motion retargeting to SMPL format, and factorized appearance control over individual clothing components. (b) Combining 3DGS scenes and collision geometry, the pipeline constructs episodes with divergent target futures and renders observations across diverse embodiments and camera configurations.}
    \label{fig:dataset_overview}
    \vspace{-1.5em}
\end{figure*}

We introduce \dataset{}, a comprehensive embodied visual tracking dataset and diagnostic benchmark designed to systematically evaluate how robots follow a designated pedestrian under perceptual ambiguity and divergent target futures. \dataset{} encompasses three core evaluation tasks: Single-Target Tracking (STT), Distracted Tracking (DT), and Branched Future Tracking (BFT). Figure~\ref{fig:dataset_overview} illustrates the data generation pipeline and avatar diversity.

\begin{table}[t]
    \centering
    \caption[Comparison of tracking benchmarks]{\textbf{Comparison of tracking benchmarks.} Platform columns indicate wheeled, quadruped, and humanoid coverage.}
    \vspace{-0.5em}
    \label{tab:benchmark_comparison}
    \scriptsize
    \setlength{\tabcolsep}{1pt}
    \renewcommand{\arraystretch}{1.1}

    \resizebox{\columnwidth}{!}{\begin{tabular*}{\columnwidth}{@{\extracolsep{\fill}}llrccccc@{}}
        \toprule
        \multirow{2}{*}{\textbf{Benchmark}} & \multirow{2}{*}{\textbf{Environment}} & \multirow{2}{*}{\shortstack{\textbf{Seq./}\\\textbf{Eps.}}} & \multirow{2}{*}{\shortstack{\textbf{Closed-}\\\textbf{loop}}} & \multirow{2}{*}{\shortstack{\textbf{Paired}\\\textbf{futures}}} & \multicolumn{3}{c}{\textbf{Platform}} \\
        & & & & & \makebox[2em][c]{\includegraphics[height=1.2em]{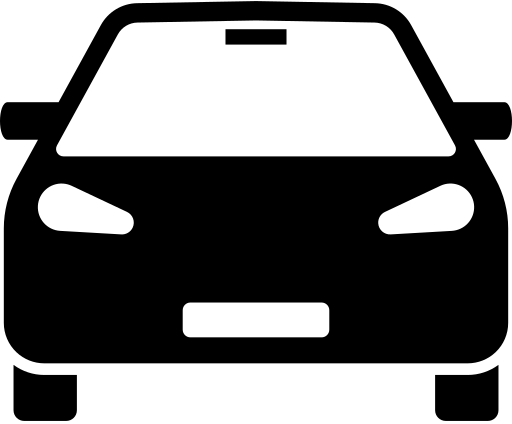}} & \makebox[2em][c]{\includegraphics[height=1.2em]{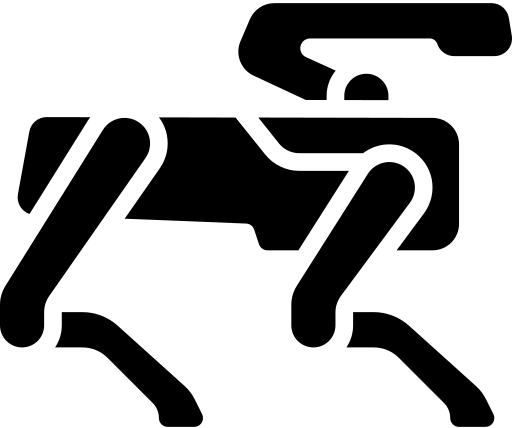}} & \makebox[2em][c]{\includegraphics[height=1.2em]{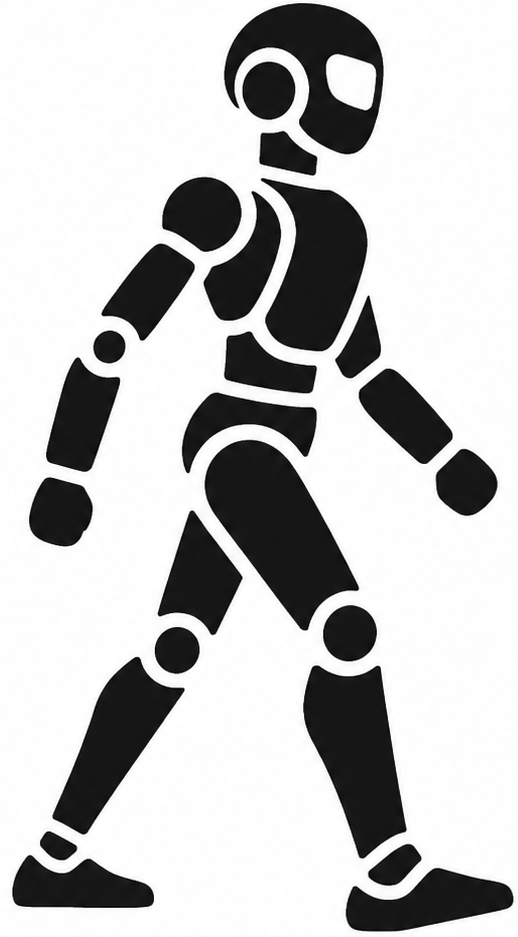}} \\
        \midrule
        PTUA~\cite{zhang2023robot} & Videos & 45 & $\times$ & $\times$ & $\checkmark$ & $\times$ & $\times$ \\
        TPT-Bench~\cite{ye2025tpt} & Videos & 48 & $\times$ & $\times$ & $\checkmark$ & $\times$ & $\times$ \\
        Follow-Bench~\cite{ye2025followbench} & IR-SIM (2D) & 1,600 & $\checkmark$ & $\times$ & $\checkmark$ & $\times$ & $\times$ \\
        OccEVT-Bench~\cite{wen2026vpevt} & Habitat-GS (3D) & 400 & $\checkmark$ & $\times$ & $\checkmark$ & $\times$ & $\times$ \\
        EVT-Bench~\cite{trackvla2025} & Habitat (3D) & 25,986 & $\checkmark$ & $\times$ & $\times$ & $\checkmark$ & $\times$ \\
        \midrule
        \textbf{\dataset{}} & \textbf{Isaac Sim (3D)} & \textbf{35,003} & $\checkmark$ & $\checkmark$ & $\checkmark$ & $\checkmark$ & $\checkmark$ \\
        \bottomrule
    \end{tabular*}}
    \vspace{-1.5em}
\end{table}

\subsection{Scene and Episode Construction}

We start from the 1,000 synthetic 3D Gaussian Splatting (3DGS) scenes in InteriorGS~\cite{sage3d} and retain 912 after filtering. In Isaac Sim, we combine 3DGS scene appearance with collision meshes for closed-loop tracking. Pedestrians follow routes sampled from the navigable regions of each scene. Our avatar assets support motion retargeting across SMPL, SMPL+H, and SMPL-X body models. Hats, tops, bottoms, and shoes can be varied independently to control pedestrian appearance. The pipeline renders egocentric RGB and depth observations for wheeled, quadrupedal, and humanoid robots under different camera configurations.

\subsection{Tracking Tasks}
Unlike prior datasets that focus solely on success rates, \dataset{} is structured to isolate specific failure modes through three progressive tasks:

\noindent\textbf{Single-Target Tracking (STT).}
Following EVT-Bench, the agent follows one person to evaluate basic tracking.

\noindent\textbf{Distracted Tracking (DT).}
To systematically evaluate policy robustness against \emph{perceptual ambiguity}, the robot must track the target amidst appearance-similar distractors. We quantify and control target--distractor appearance similarity using graduated difficulty levels based on the 1-Wasserstein distance in CIELAB space (Lab-W1)~\cite{rubner2000emd}. This enables fine-grained diagnosis of how visual confusions and partial occlusions impact tracking stability.

\noindent\textbf{Branched Future Tracking (BFT).}
To evaluate policy resilience against \emph{planning uncertainty} under divergent target motions, we introduce a counterfactual evaluation protocol. Episode groups are constructed to share an identical observation--action history up to a branching point, after which the target diverges into alternative future trajectories (e.g., continuing straight versus turning left or right). By evaluating paired closed-loop rollouts from the same historical context, BFT measures whether a tracking policy can successfully adapt to counterfactual future intents.

\subsection{Dataset Statistics}

\dataset{} comprises 35,003 rigorous episodes spanning three robot platforms: the Unitree Go2 quadruped, the Dingo wheeled base, and the Unitree G1 humanoid. We split the dataset by scene, with 34,103 training episodes from 837 scenes and 900 test episodes from 75 held-out scenes. The test split contains 300 episodes per task.

Table~\ref{tab:benchmark_comparison} provides a structural comparison between \dataset{} and existing tracking benchmarks, highlighting our unique support for factorized appearance control, counterfactual future branching, and multi-embodiment closed-loop evaluation.

\section{Experiments}
\label{sec:experiments}

\begin{table*}[!t]
\centering
\caption{\textbf{Performance on EVT-Bench.}
\method{} achieves the state-of-the-art performance across all three tasks.
\textbf{Bold} marks the best reported value in each column. $\dagger$ uses Grounding DINO~\cite{liu2024groundingdino}. Dashes (--) denote unavailable baseline results. Params reports model scale.}
\vspace{-0.5em}
\label{tab:evtbench_results}
\footnotesize
\setlength{\tabcolsep}{2.2pt}
\renewcommand{\arraystretch}{0.94}
\begin{tabularx}{\textwidth}{@{}llc*{9}{>{\centering\arraybackslash}X}@{}}
\toprule
\multirow{2}{*}{Method} & \multirow{2}{*}{Venue} & \multirow{2}{*}{Params}
  & \multicolumn{3}{c}{Single-Target Tracking (STT)}
  & \multicolumn{3}{c}{Distracted Tracking (DT)}
  & \multicolumn{3}{c}{Ambiguity Tracking (AT)} \\
\cmidrule(lr){4-6}\cmidrule(lr){7-9}\cmidrule(l){10-12}
& & & SR$\uparrow$ & TR$\uparrow$ & CR$\downarrow$
& SR$\uparrow$ & TR$\uparrow$ & CR$\downarrow$
& SR$\uparrow$ & TR$\uparrow$ & CR$\downarrow$ \\
\midrule
\rowcolor{black!8}[0pt][0pt]
\multicolumn{12}{@{}l@{}}{\textit{RL-based policies}} \\
PoliFormer$^{\dagger}$~\cite{zeng2024poliformer} & CoRL '24 & $<$1B & 4.7 & 15.5 & 40.1 & 2.6 & 13.2 & 44.5 & 3.0 & 15.4 & 41.5 \\
EVT~\cite{zhong2024offlineevt} & ECCV '24 & $<$1B & 24.4 & 39.1 & 42.5 & 3.2 & 11.2 & 47.9 & 17.4 & 21.1 & 45.6 \\
\midrule
\rowcolor{black!8}[0pt][0pt]
\multicolumn{12}{@{}l@{}}{\textit{Modular pipelines}} \\
ORTrack~\cite{wu2025ortrack}+PID~\cite{astrom2006pid} & CVPR '25 & $<$1B & 31.9 & 46.6 & 4.6 & 19.2 & 42.0 & \textbf{4.6} & 16.9 & 38.4 & \textbf{5.2} \\
OA-VAT~\cite{oavat2026} & CVPR '26 & $<$1B & 33.2 & 49.3 & 6.9 & 17.4 & 40.1 & 8.3 & 17.3 & 39.6 & 7.3 \\
\midrule
\rowcolor{black!8}[0pt][0pt]
\multicolumn{12}{@{}l@{}}{\textit{MLLM-based methods}} \\
Uni-NaVid~\cite{zhang2024uninavid} & RSS '25 & 7B & 25.7 & 39.5 & 41.9 & 11.3 & 27.4 & 43.5 & 8.3 & 28.6 & 43.7 \\
NavFoM~\cite{zhang2025navfom} & ICLR '26 & 7B & 85.0 & 80.5 & -- & 61.4 & 68.2 & -- & -- & -- & -- \\
TrackVLA~\cite{trackvla2025} & CoRL '25 & 7B & 85.1 & 78.6 & 1.7 & 57.6 & 63.2 & 5.8 & 50.2 & 63.7 & 17.1 \\
TrackVLA++~\cite{trackvlapp2025} & ICRA '26 & 7B & 86.0 & 81.0 & 2.1 & 66.5 & 68.8 & 4.7 & 51.2 & 63.4 & 15.9 \\
OmTrackVLA~\cite{lee2026vlxgorobustembodiedtracking} & Tech. Rep. '26 & $\sim$1B & 81.4 & 82.8 & 5.1 & 41.5 & 58.8 & 11.3 & 60.0 & 73.9 & 7.6 \\
MiniCPM-RobotTrack~\cite{minicpmrobot2026} & Tech. Rep. '26 & $<$1B & 84.1 & \textbf{89.8} & 3.0 & 53.2 & 73.4 & 13.6 & 58.0 & 80.4 & 9.0 \\
Qwen-RobotNav~\cite{qwenrobotnav2026} & Tech. Rep. '26 & 8B & 78.6 & 89.7 & 5.7 & -- & -- & -- & -- & -- & -- \\
ABot-N0~\cite{abotn02026} & Tech. Rep. '26 & 4B & 86.9 & 87.6 & 8.5 & 66.7 & 75.4 & 11.6 & 67.3 & 79.5 & 7.1 \\
LightNav-0~\cite{lightnav2026} & Tech. Rep. '26 & 4B & 91.7 & 87.7 & 1.9 & 82.6 & \textbf{80.1} & \textbf{4.6} & -- & -- & -- \\
\midrule
\rowcolor{cvprblue!12}
\method{} & Ours & $<$1B & \textbf{92.2} & 89.7 & \textbf{0.7} & \textbf{84.3} & \textbf{80.1} & 5.2 & \textbf{76.1} & \textbf{85.8} & \textbf{4.1}\\
\bottomrule
\end{tabularx}
\vspace{-1.5em}
\end{table*}

\begin{table}[t]
\centering
\caption{\textbf{STT performance and efficiency on \dataset{}.} \method{} outperforms all open-source baselines in STT while achieving lower latency than all MLLM-based methods.}
\vspace{-0.5em}
\label{tab:forktrack_results}
\footnotesize
\setlength{\tabcolsep}{1.25pt}
\begin{tabularx}{\columnwidth}{@{}l*{3}{>{\centering\arraybackslash}X}ccc@{}}
\toprule
\multirow{2}{*}{Method} & \multicolumn{3}{c}{STT}
  & \multicolumn{3}{c@{}}{Inference efficiency} \\
\cmidrule(lr){2-4}\cmidrule(l){5-7}
& SR$\uparrow$ & TR$\uparrow$ & CR$\downarrow$
& {\scriptsize Params\,(M)$\!\downarrow$}
& {\scriptsize MACs\,(G/step)$\!\downarrow$}
& {\scriptsize Lat.\,(ms)$\!\downarrow$} \\
\midrule
\rowcolor{black!8}[0pt][0pt]
\multicolumn{7}{@{}l@{}}{\textit{RL-based policies}} \\
PoliFormer$^{\dagger}$~\cite{zeng2024poliformer} & 1.3 & 53.4 & 56.3 & 215.2 & 449.9 & 173.7 \\
EVT~\cite{zhong2024offlineevt} & 0.7 & 75.1 & 71.7 & 883.8 & 237.9 & 48.3 \\
\midrule
\rowcolor{black!8}[0pt][0pt]
\multicolumn{7}{@{}l@{}}{\textit{Modular pipelines}} \\
ORTrack+PID~\cite{wu2025ortrack,astrom2006pid} & 38.7 & 81.0 & 15.0 & \textbf{8.1} & \textbf{2.9} & \textbf{6.8} \\
OA-VAT~\cite{oavat2026} & 38.3 & 80.4 & 12.7 & 402.6 & 8.5 & 10.9 \\
\midrule
\rowcolor{black!8}[0pt][0pt]
\multicolumn{7}{@{}l@{}}{\textit{MLLM-based methods}} \\
Uni-NaVid~\cite{zhang2024uninavid} & 27.7 & 78.7 & 20.3 & 7746.9 & 2583.7 & 175.1 \\
OmTrackVLA~\cite{lee2026vlxgorobustembodiedtracking} & 46.0 & 85.0 & 13.7 & 1059.1 & 457.7 & 98.0 \\
MiniCPM~\cite{minicpmrobot2026} & 36.3 & 82.6 & 22.3 & 899.6 & 436.0 & 94.3 \\
LightNav-0~\cite{lightnav2026} & 28.7 & 81.2 & 29.7 & 4448.3 & 4820.1 & 344.0 \\
\midrule
\rowcolor{cvprblue!12}
\textbf{\method{}} & \textbf{49.0} & \textbf{87.4} & \textbf{6.1} & 367.0 & 50.7 & 68.2 \\
\bottomrule
\end{tabularx}
\vspace{-1.5em}
\end{table}

\subsection{Experimental Setup}
\label{sec:experimental_setup}

\noindent\textbf{Baselines.} We group the compared methods by their primary action-generation mechanism. RL-based policies include PoliFormer~\cite{zeng2024poliformer} and EVT~\cite{zhong2024offlineevt}. Modular pipelines include ORTrack~\cite{wu2025ortrack}+PID~\cite{astrom2006pid} and OA-VAT~\cite{oavat2026}. MLLM-based methods include Uni-NaVid~\cite{zhang2024uninavid}, NavFoM~\cite{zhang2025navfom}, TrackVLA~\cite{trackvla2025}, TrackVLA++~\cite{trackvlapp2025}, OmTrackVLA~\cite{lee2026vlxgorobustembodiedtracking}, MiniCPM-RobotTrack (MiniCPM)~\cite{minicpmrobot2026}, Qwen-RobotNav~\cite{qwenrobotnav2026}, ABot-N0~\cite{abotn02026}, and LightNav-0~\cite{lightnav2026}.

\noindent\textbf{Benchmarks.} We evaluate STT, DT, and Ambiguity Tracking (AT) on EVT-Bench~\cite{trackvla2025} and STT, DT, and BFT on \dataset{}. On \dataset{}, we use a Unitree Go2 with the evaluation criteria and metrics of the EVT-Bench Spot setup. More details are in the supplementary material. Real-world trials use a Unitree Go2 quadruped robot. For all main benchmark results, our method uses the vision-only configuration.

\noindent\textbf{Metrics.} For closed-loop tracking, we report success rate (SR), tracking rate (TR), and collision rate (CR) as defined in EVT-Bench. All three rates are reported as percentages, with percent signs omitted in the tables. For inference efficiency, we report the number of online model parameters (Params, M), multiply--accumulate operations per step (MACs, G/step), and per-frame latency (Lat., ms). For real-robot experiments, we report average (Avg.) and 95th-percentile (P95) inference latency in milliseconds.

\subsection{Simulation Benchmark Results}
\label{sec:sim_results}

\noindent\textbf{Performance on EVT-Bench.}
As shown in Table~\ref{tab:evtbench_results}, we first evaluate our method on the mainstream EVT-Bench benchmark~\cite{trackvla2025}. \method{} achieves the highest reported success rate on all three EVT-Bench tasks, with 92.2\% on STT, 84.3\% on DT, and 76.1\% on AT. On DT, it surpasses LightNav-0~\cite{lightnav2026} by 1.7 percentage points in SR. On AT, it improves SR from 67.3\% with ABot-N0~\cite{abotn02026} to 76.1\% and achieves the highest TR (85.8\%) and lowest CR (4.1\%). \method{} also records the lowest CR on STT (0.7\%). Notably, with fewer than 1B parameters, \method{} exceeds the best reported SR among the MLLM-based baselines on each task, offering a compact solution for deployment.

Beyond our method, the results reveal a clear performance hierarchy across these action-generation paradigms: MLLM-based methods outperform modular pipelines, which in turn surpass RL-based policies. On DT, the best SR in each group is 82.6\%, 19.2\%, and 3.2\%, respectively. Modular ORTrack~\cite{wu2025ortrack}+PID~\cite{astrom2006pid} matches the lowest DT collision rate (4.6\%) but reaches only 42.0\% TR, indicating that low collisions largely stem from target loss rather than effective obstacle avoidance, and underscoring the need to tightly couple target lock with reactive navigation.

\begin{figure}[t]
    \centering
    \includegraphics[width=\linewidth]{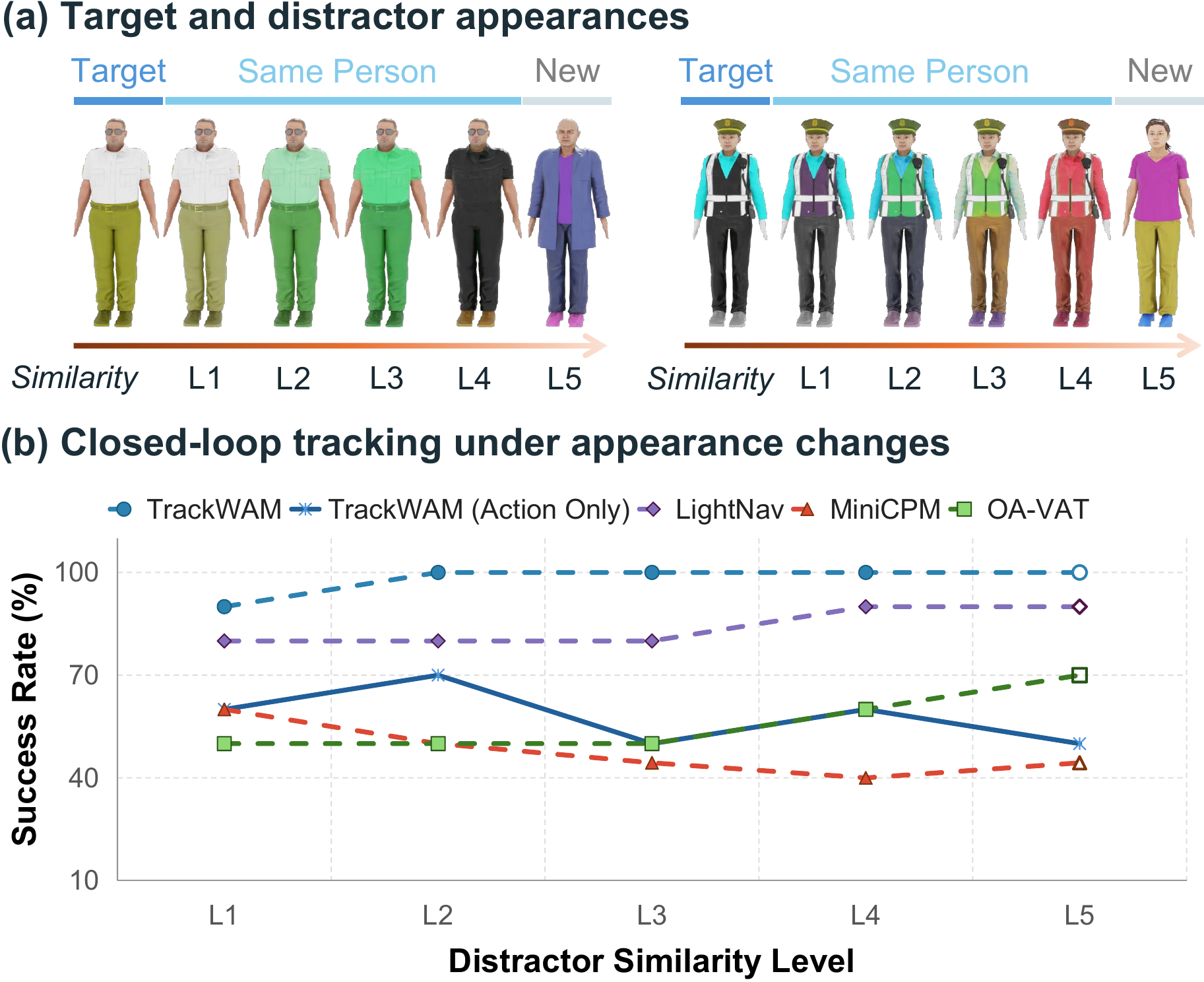}
    \caption{\textbf{Distracted tracking across similarity levels on \dataset{}.} Across five similarity levels, we observe that for methods with explicit perceptual supervision, tracking success rises as distractors become visually different.}
    \label{fig:forktrack_dt_similarity}
    \vspace{-1.5em}
\end{figure}

\begin{figure}[t]
    \centering
    \includegraphics[width=\linewidth]{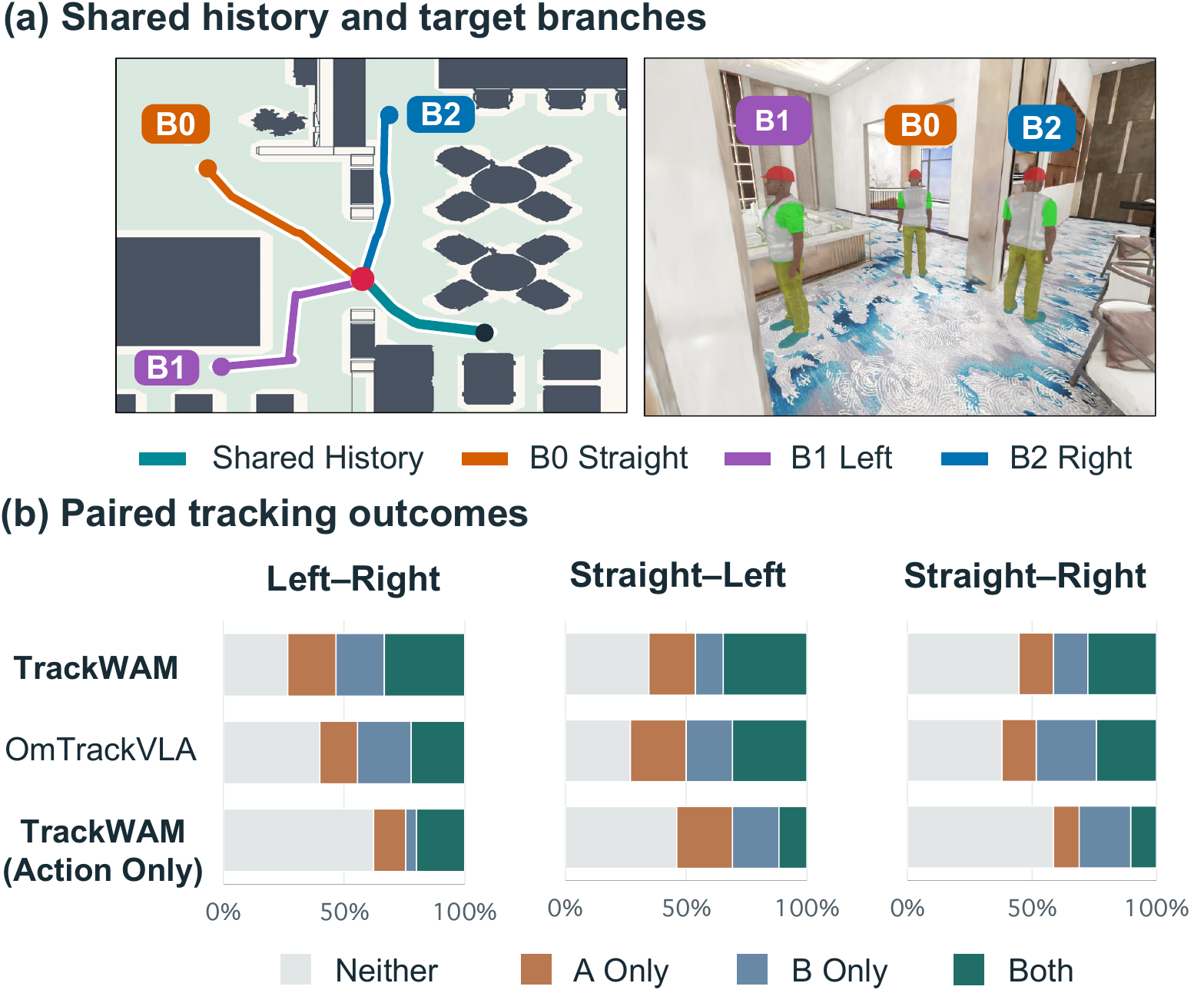}
    \caption{\textbf{Counterfactual branching evaluation on \dataset{}.} Paired tracking outcomes showing that world expert enables reliable counterfactual reasoning across alternative futures.}
    \label{fig:forktrack_bft_postbranch}
    \vspace{-1em}
\end{figure}

\noindent\textbf{Performance and Efficiency on \dataset{}.}
We evaluate all methods on the unseen scenes of \dataset{}. In Table~\ref{tab:forktrack_results}, \method{} achieves the highest STT success rate (49.0\%) and tracking rate (87.4\%), and the lowest collision rate (6.1\%). It improves SR by 3.0 percentage points over OmTrackVLA, the strongest baseline by SR. The supplementary material reports quantitative results for DT and BFT. Figure~\ref{fig:forktrack_dt_similarity} analyzes DT across levels of target--distractor appearance similarity, and Fig.~\ref{fig:forktrack_bft_postbranch} compares BFT outcomes across paired target trajectories.

Besides, we also restrict this comparison to open-source baselines so that we can run and profile each method under the same hardware setup. We measure batch-1 inference per step on the same NVIDIA RTX 4090, with each method's native preprocessing and numerical precision. \method{} uses fewer parameters and MACs than each evaluated MLLM-based baseline. It also runs faster, at 68.2~ms per frame versus 94.3~ms for the fastest MLLM-based baseline. The modular pipelines use fewer MACs and run faster, while their SRs are at least 10.3 percentage points lower. Both RL-based policies also achieve lower performance. Because some EVT-Bench baselines lack public implementations, we summarize the available throughput and latency figures in the supplementary material.

\noindent\textbf{Sensitivity to Distractor Appearance Similarity.}
To isolate visual ambiguity, we evaluate two \dataset{} scenes featuring five distractor levels ranked from closest (L1) to farthest (L5) by Lab-W1 distance (exact values in the supplementary material). As shown in Fig.~\ref{fig:forktrack_dt_similarity}, methods with explicit perceptual supervision (OA-VAT, LightNav-0, and \method{}) improve monotonically or stay stable as distractors become more distinct, with \method{} achieving the highest overall robustness. Conversely, MiniCPM and Action Only fluctuate inconsistently across similarity levels, confirming the necessity of explicit perceptual supervision to disambiguate similar distractors.

\noindent\textbf{Adaptability to divergent target futures.}
The BFT split of \dataset{} tests policies through counterfactual evaluation. Figure~\ref{fig:forktrack_bft_postbranch} compares tracking outcomes after different continuations from a shared observation--action history. OmTrackVLA succeeds more often on single branches in straight-inclusive pairings, due to a higher concentration of successful straight trajectories. In contrast, \method{} achieves the highest rate of both-branch success across all three pairings and consistently outperforms its Action-Only variant. This demonstrates that our world expert enables robust counterfactual reasoning.

\subsection{Ablation Studies}
\label{sec:ablations}

\begin{table}[t]
    \centering
    \caption{\textbf{Ablation Study of Proposed Designs.} We analyze the contributions of individual components on EVT-Bench STT split.}
    \vspace{-0.5em}
    \label{tab:component_ablation}
    \footnotesize
    \setlength{\tabcolsep}{2pt}
    \begin{tabularx}{\columnwidth}{@{}>{\raggedright\arraybackslash}X*{3}{>{\centering\arraybackslash}p{0.12\columnwidth}}@{}}
        \toprule
        Variant & SR$\uparrow$ & TR$\uparrow$ & CR$\downarrow$ \\
        \midrule
        \rowcolor{black!8}[0pt][0pt]
        \multicolumn{4}{@{}l@{}}{\textit{Prediction-to-action comparison}} \\
        \method{} (Action Only) & 63.1 & 67.8 & 19.1 \\
        Parallel prediction & 89.4 & 85.9 & 2.1 \\
        \rowcolor{cvprblue!12}
        \textbf{\method{} (Full)} & 92.2 & \textbf{89.7} & \textbf{0.7} \\
        \midrule
        \rowcolor{black!8}[0pt][0pt]
        \multicolumn{4}{@{}l@{}}{\textit{Training ablations}} \\
        w/o spatial supervision & 65.8 & 84.2 & 1.9 \\
        w/o future-feature conditioning & 85.8 & 84.0 & 2.2 \\
        Copy-last feature conditioning & 85.7 & 83.7 & 1.9 \\
        \midrule
        \rowcolor{black!8}[0pt][0pt]
        \multicolumn{4}{@{}l@{}}{\textit{Inference ablations}} \\
        Current frame only & 66.1 & 73.7 & 2.2 \\
        One-way fusion & 64.7 & 65.1 & 9.0 \\
        \midrule
        \rowcolor{black!8}[0pt][0pt]
        \multicolumn{4}{@{}l@{}}{\textit{Additional sensor input}} \\
        \method{} (+ UWB input) & \textbf{92.5} & 88.6 & 2.4 \\
        \bottomrule
    \end{tabularx}
    \vspace{-1.5em}
\end{table}

Table~\ref{tab:component_ablation} presents ablations on the STT split of EVT-Bench.

\noindent\textbf{Predictive conditioning.} Adding target-state prediction raises SR from 63.1\% to 89.4\%, and feeding it into the control loop further increases SR to 92.2\% while cutting CR from 2.1\% to 0.7\%, demonstrating its effectiveness in both training supervision and action conditioning.

\noindent\textbf{Training supervision.} Removing spatial supervision drops SR to 65.8\%, showing the necessity of geometric guidance. Meanwhile, removing future-feature conditioning or replacing it with copy-last features yields comparable SR (85.8\% and 85.7\%); we retain the frozen predictor to stabilize long-horizon predictions under occlusions.

\noindent\textbf{Temporal context and fusion.} Using only the current frame drops SR from 92.2\% to 66.1\%, while replacing Track Perceiver Blocks with one-way fusion reduces SR to 64.7\% and increases CR from 0.7\% to 9.0\%, confirming the need for both historical context and bidirectional fusion.

\noindent\textbf{Optional UWB modality.} While our primary model operates purely on vision, \method{} supports optional UWB coordinates for real-world deployment. Adding UWB further raises STT success rate to 92.5\% via reliable global guidance. The slight trade-offs, namely a minor decrease in TR (89.7\% to 88.6\%) and an increase in CR (0.7\% to 2.4\%), stem from more aggressive, direct-to-target paths that occasionally reduce reactive obstacle avoidance.

\subsection{Real-World Deployment}
\label{sec:real_robot}

\begin{table}[t]
\centering
\caption{\textbf{Real-world tracking performance and inference latency on a robot.}
\method{} achieves the highest success rate and the lowest inference latency among the evaluated methods.}
\vspace{-0.5em}
\label{tab:real_robot_results}
\footnotesize
\setlength{\tabcolsep}{2.6pt}
\begin{tabularx}{\columnwidth}{@{}>{\raggedright\arraybackslash}Xccccccc@{}}
\toprule
\multirow{2}{*}{Method} & \multicolumn{3}{c}{Success by scenario}
& \multicolumn{2}{c}{Overall (\%)}
& \multicolumn{2}{c@{}}{Inference (ms)} \\
\cmidrule(lr){2-4}\cmidrule(lr){5-6}\cmidrule(l){7-8}
& Indoor & Walkway & Crowd & SR$\uparrow$ & TR$\uparrow$ & Avg.$\downarrow$ & P95$\downarrow$ \\
\midrule
LightNav-0 & 1/5 & 0/5 & 0/5 & 6.7 & 71.6 & 357.2 & 435.0 \\
MiniCPM & \textbf{4/5} & 3/5 & 1/5 & 53.3 & 82.1 & 210.3 & 215.3 \\
OmTrackVLA & 3/5 & 2/5 & 0/5 & 33.3 & 78.9 & 120.8 & 126.4 \\
\rowcolor{cvprblue!12}
\method{} & \textbf{4/5} & \textbf{5/5} & \textbf{3/5} & \textbf{80.0} & \textbf{97.2} & \textbf{100.2} & \textbf{103.9} \\
\bottomrule
\end{tabularx}
\end{table}

\begin{figure}[t]
    \centering
    \includegraphics[width=0.90\linewidth]{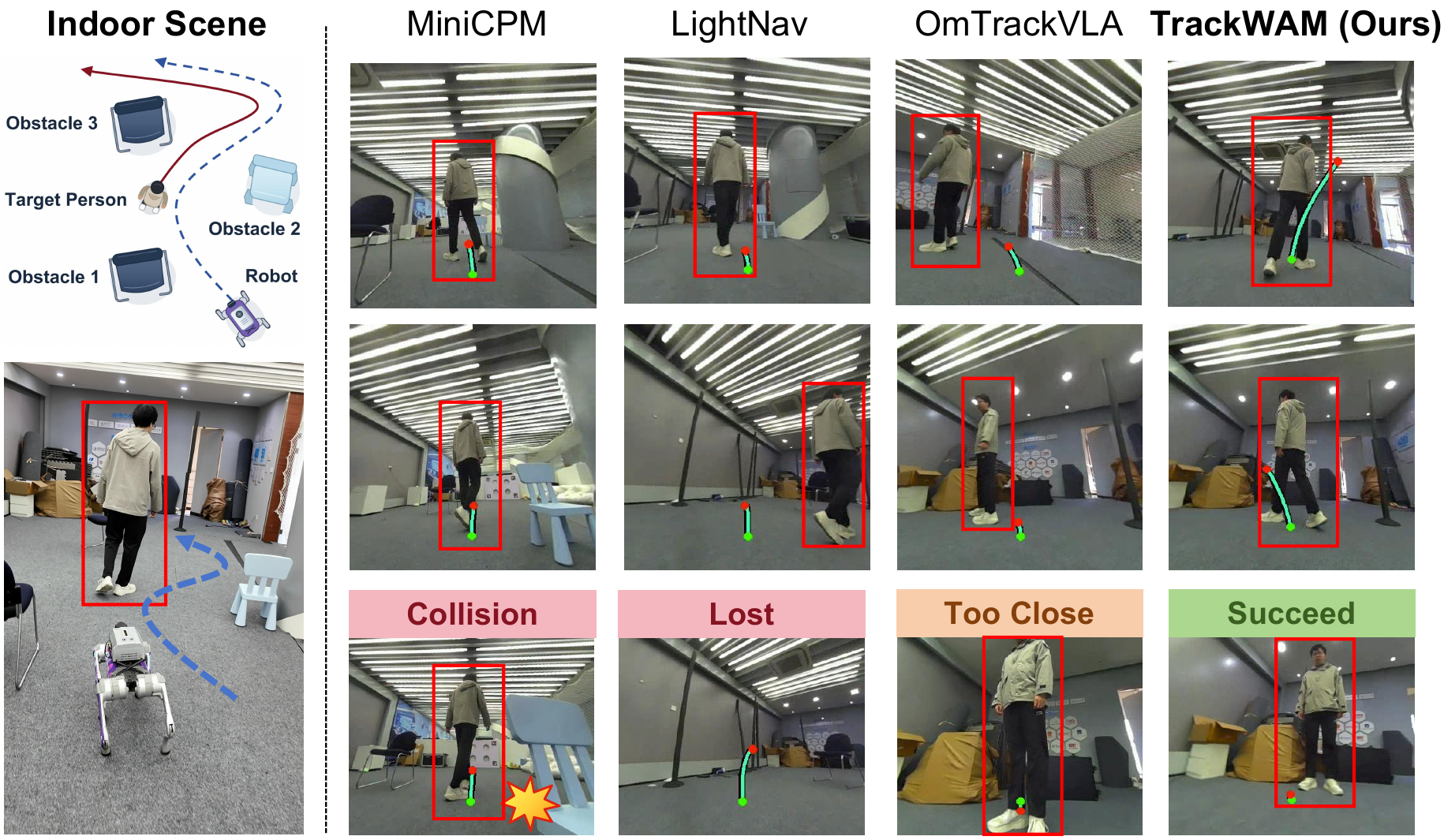}
    \caption{\textbf{Qualitative analysis of indoor real-world tracking.}
\method{} completes the route by avoiding obstacles, keeping pace with the target, and stopping at a safe distance.}
    \label{fig:real_robot_indoor}
    \vspace{-1.5em}
\end{figure}

\noindent\textbf{Evaluation setup.} We deploy \method{} and the baselines on a Unitree Go2 quadruped robot to validate real-world tracking performance and inference latency. Given the cost of physical trials, we select three representative MLLM-based baselines. OmTrackVLA~\cite{lee2026vlxgorobustembodiedtracking} and MiniCPM-RobotTrack~\cite{minicpmrobot2026} represent compact models, while LightNav-0~\cite{lightnav2026} achieves the highest reported SR among baselines on EVT-Bench.

\noindent\textbf{Quantitative results.} In Table~\ref{tab:real_robot_results}, \method{} achieves the highest overall SR (80.0\%) and TR (97.2\%). It ties with MiniCPM indoors (4/5) and succeeds more often on the walkway (5/5 vs.\ 3/5) and in crowds (3/5 vs.\ 1/5), yielding a 26.7-point gain in overall SR. \method{} also achieves the lowest average and P95 inference latencies (100.2 and 103.9~ms), reducing them by 17.1\% and 17.8\%, respectively, compared with OmTrackVLA.
This shows \method{} balances tracking accuracy and low latency.

\begin{figure}[t]
    \centering
    \includegraphics[width=\linewidth]{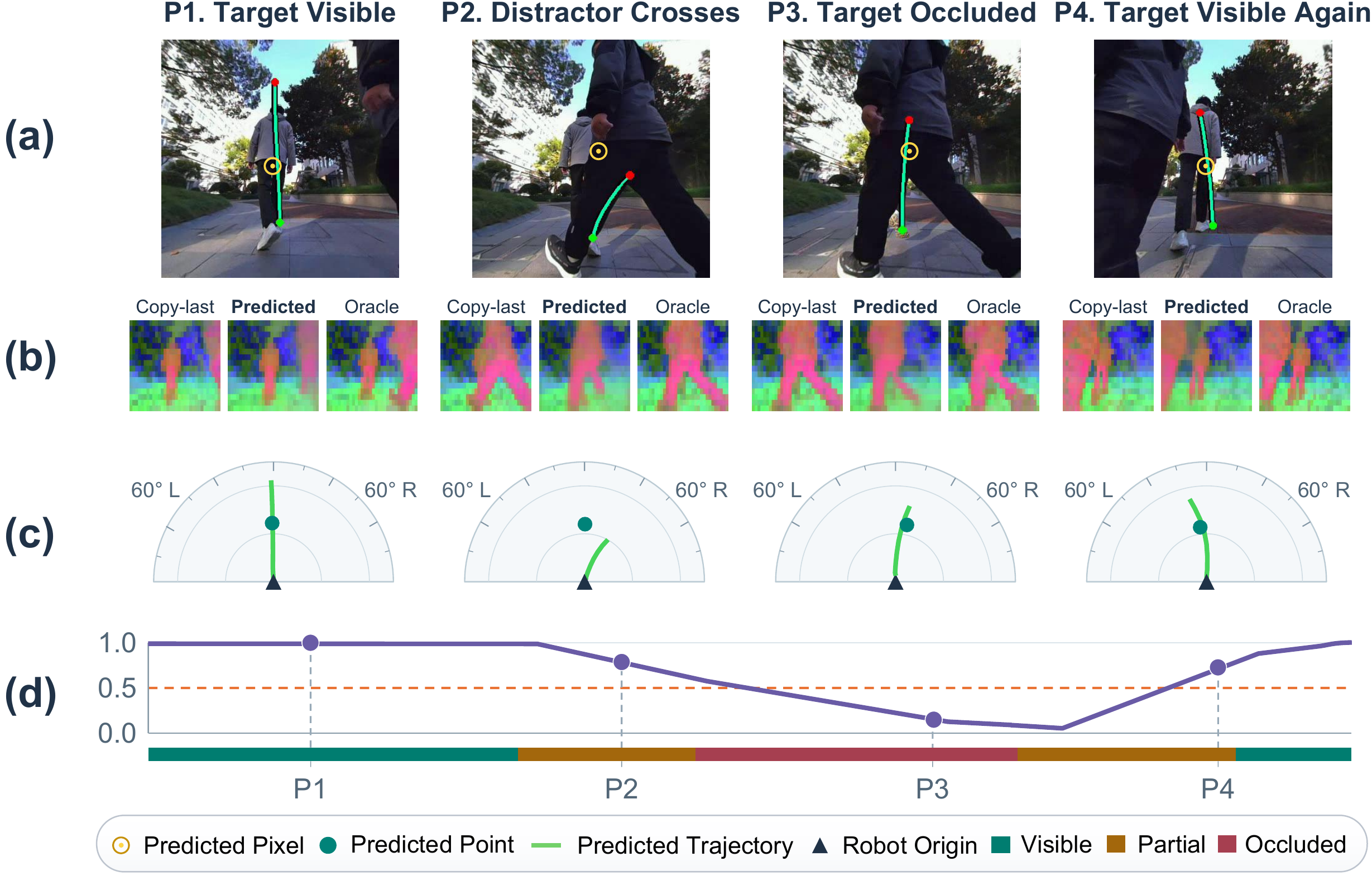}
    \caption{\textbf{Real-robot tracking through occlusion.}
(a) egocentric RGB with tracking overlays, (b) future DINO features, (c) robot-frame target estimates and trajectories, and (d) predicted target visibility. The visibility intervals are manually annotated.}
    \label{fig:real_robot_diagnostics}
    \vspace{-1.5em}
\end{figure}

\noindent\textbf{Qualitative Analysis.}
We further analyze representative trials to explore distinct failure modes across methods. As shown in Fig.~\ref{fig:real_robot_indoor}, the indoor route features narrow passages, S-shaped turns, and occluding obstacles. MiniCPM collides with a chair due to insufficient clearance, while LightNav-0 loses the target as high latency limits timely reactions. OmTrackVLA completes the route but approaches the target too closely upon stopping. In contrast, \method{} avoids obstacles while maintaining a safe following distance.

Figure~\ref{fig:real_robot_diagnostics} shows how \method{} maintains tracking through a temporary occlusion in egocentric view. As a pedestrian distractor crosses between the target and the agent, the predicted future DINO features capture the distractor's leftward motion. During the occlusion, the spatial model continues to estimate the target's visibility, image-plane position, and relative position in the robot frame, with visibility responding after a slight delay. Then the policy generates an effective tracking trajectory. Once the distractor moves clear, the robot promptly reacquires the target.

\section{Conclusion}
We present \method{}, a lightweight embodied tracking framework that couples spatial-perceptual supervision with closed-loop target-state prediction, and \dataset{}, a benchmark for diagnosing embodied tracking under visual ambiguity and divergent target futures. Without heavy MLLMs, \method{} leverages frozen visual foundation features, future-feature conditioning, and bidirectional cross-attention to handle ambiguity, occlusions, and counterfactual trajectory branches. Experiments on EVT-Bench and \dataset{} show higher tracking accuracy and lower inference latency than evaluated MLLM-based baselines. Deployment on a Unitree Go2 robot further validates its robustness in complex indoor, outdoor, and crowded scenes.
We hope our work supports future research in EVT.

{\small
\bibliographystyle{ieeenat_fullname}
\bibliography{evt_references}

\begin{thebibliography}{54}
\providecommand{\natexlab}[1]{#1}
\providecommand{\url}[1]{\texttt{#1}}
\expandafter\ifx\csname urlstyle\endcsname\relax
  \providecommand{\doi}[1]{doi: #1}\else
  \providecommand{\doi}{doi: \begingroup \urlstyle{rm}\Url}\fi

\bibitem[{AMAP CV Lab}(2026)]{abotn02026}
{AMAP CV Lab}.
\newblock {ABot-N0}: Technical report on the {VLA} foundation model for
  versatile embodied navigation.
\newblock \href{https://arxiv.org/abs/2602.11598}{arXiv:2602.11598}, 2026.

\bibitem[{\AA}str{\"o}m and H{\"a}gglund(2006)]{astrom2006pid}
Karl~Johan {\AA}str{\"o}m and Tore H{\"a}gglund.
\newblock \emph{Advanced {PID} Control}.
\newblock ISA -- The Instrumentation, Systems, and Automation Society, 2006.

\bibitem[Babenko et~al.(2009)Babenko, Yang, and Belongie]{babenko2009visual}
Boris Babenko, Ming-Hsuan Yang, and Serge Belongie.
\newblock Visual tracking with online multiple instance learning.
\newblock In \emph{Proceedings of the IEEE Conference on Computer Vision and
  Pattern Recognition}, pages 983--990, 2009.

\bibitem[Bai et~al.(2025)Bai, Cai, Chen, et~al.]{bai2025qwen3vl}
Shuai Bai, Yuxuan Cai, Ruizhe Chen, et~al.
\newblock {Qwen3-VL} technical report.
\newblock \href{https://arxiv.org/abs/2511.21631}{arXiv:2511.21631}, 2025.

\bibitem[Bar et~al.(2025)Bar, Zhou, Tran, Darrell, and LeCun]{bar2025nwm}
Amir Bar, Gaoyue Zhou, Danny Tran, Trevor Darrell, and Yann LeCun.
\newblock Navigation world models.
\newblock In \emph{Proceedings of the IEEE/CVF Conference on Computer Vision
  and Pattern Recognition}, 2025.

\bibitem[Darcet et~al.(2024)Darcet, Oquab, Mairal, and
  Bojanowski]{darcet2024registers}
Timoth{\'e}e Darcet, Maxime Oquab, Julien Mairal, and Piotr Bojanowski.
\newblock Vision transformers need registers.
\newblock In \emph{International Conference on Learning Representations}, 2024.

\bibitem[Everingham et~al.(2010)Everingham, Van~Gool, Williams, Winn, and
  Zisserman]{everingham2010pascal}
Mark Everingham, Luc Van~Gool, Christopher~KI Williams, John Winn, and Andrew
  Zisserman.
\newblock The {PASCAL} visual object classes ({VOC}) challenge.
\newblock \emph{International journal of computer vision}, 88\penalty0
  (2):\penalty0 303--338, 2010.

\bibitem[Fan et~al.(2019)Fan, Lin, Yang, Chu, Deng, Yu, Bai, Xu, Liao, and
  Ling]{fan2019lasot}
Heng Fan, Liting Lin, Fan Yang, Peng Chu, Ge Deng, Sijia Yu, Hexin Bai, Yong
  Xu, Chunyuan Liao, and Haibin Ling.
\newblock {LaSOT}: A high-quality benchmark for large-scale single object
  tracking.
\newblock In \emph{Proceedings of the IEEE/CVF Conference on Computer Vision
  and Pattern Recognition (CVPR)}, pages 5374--5383, 2019.

\bibitem[Ha and Schmidhuber(2018)]{ha2018worldmodels}
David Ha and J{\"u}rgen Schmidhuber.
\newblock Recurrent world models facilitate policy evolution.
\newblock In \emph{Advances in Neural Information Processing Systems 31}, pages
  2451--2463. Curran Associates, Inc., 2018.

\bibitem[Hafner et~al.(2025)Hafner, Pasukonis, Ba, and
  Lillicrap]{hafner2025dreamerv3}
Danijar Hafner, Jurgis Pasukonis, Jimmy Ba, and Timothy Lillicrap.
\newblock Mastering diverse control tasks through world models.
\newblock \emph{Nature}, 640:\penalty0 647--653, 2025.

\bibitem[Hansen et~al.(2024)Hansen, Su, and Wang]{hansen2024tdmpc2}
Nicklas Hansen, Hao Su, and Xiaolong Wang.
\newblock {TD-MPC2}: Scalable, robust world models for continuous control.
\newblock In \emph{International Conference on Learning Representations}, 2024.

\bibitem[Jaegle et~al.(2021)Jaegle, Gimeno, Brock, Vinyals, Zisserman, and
  Carreira]{jaegle2021perceiver}
Andrew Jaegle, Felix Gimeno, Andy Brock, Oriol Vinyals, Andrew Zisserman, and
  Joao Carreira.
\newblock Perceiver: General perception with iterative attention.
\newblock In \emph{Proceedings of the 38th International Conference on Machine
  Learning}, pages 4651--4664. PMLR, 2021.

\bibitem[Karypidis et~al.(2025)Karypidis, Kakogeorgiou, Gidaris, and
  Komodakis]{karypidis2025dinoforesight}
Efstathios Karypidis, Ioannis Kakogeorgiou, Spyridon Gidaris, and Nikos
  Komodakis.
\newblock {DINO-Foresight}: Looking into the future with {DINO}.
\newblock In \emph{Advances in Neural Information Processing Systems}, 2025.

\bibitem[Koh et~al.(2021)Koh, Lee, Yang, Baldridge, and
  Anderson]{koh2021pathdreamer}
Jing~Yu Koh, Honglak Lee, Yinfei Yang, Jason Baldridge, and Peter Anderson.
\newblock {Pathdreamer}: A world model for indoor navigation.
\newblock In \emph{Proceedings of the IEEE/CVF International Conference on
  Computer Vision}, 2021.

\bibitem[Lee et~al.(2026)Lee, Ying, and
  Zhao]{lee2026vlxgorobustembodiedtracking}
Kyusong Lee, Heting Ying, and Tiancheng Zhao.
\newblock {VLX-Go}: Robust embodied tracking through multi-stage learning and
  on-policy refinement.
\newblock \href{https://arxiv.org/abs/2610.19361}{arXiv:2610.19361}, 2026.

\bibitem[{LightNav-0 Contributors}(2026)]{lightnavthor2026}
{LightNav-0 Contributors}.
\newblock Serving {LightNav-0} on {NVIDIA Jetson AGX Thor}.
\newblock
  \href{https://github.com/lightorigins/LightNav-0/blob/main/docs/JETSON_THOR.md\#measured-throughput}{Official
  deployment documentation}, 2026.

\bibitem[Lipman et~al.(2023)Lipman, Chen, Ben-Hamu, Nickel, and
  Le]{lipman2023flow}
Yaron Lipman, Ricky T.~Q. Chen, Heli Ben-Hamu, Maximilian Nickel, and Matt Le.
\newblock Flow matching for generative modeling.
\newblock In \emph{International Conference on Learning Representations}, 2023.

\bibitem[Liu et~al.(2026{\natexlab{a}})Liu, Xie, Luo, Chu, Hu, Wu, and
  Xu]{liu2026navforesee}
Fei Liu, Shichao Xie, Minghua Luo, Zedong Chu, Junjun Hu, Xiaolong Wu, and Mu
  Xu.
\newblock {NavForesee}: A unified vision-language world model for hierarchical
  planning and dual-horizon navigation prediction.
\newblock In \emph{Proceedings of the IEEE/CVF Conference on Computer Vision
  and Pattern Recognition}, pages 32431--32440, 2026{\natexlab{a}}.

\bibitem[Liu et~al.(2026{\natexlab{b}})Liu, Qi, Zhang, Li, Wang, Wu, Ye, Zhang,
  Chen, Zhong, Zhang, and Wang]{trackvlapp2025}
Jiahang Liu, Yunpeng Qi, Jiazhao Zhang, Minghan Li, Shaoan Wang, Kui Wu,
  Hanjing Ye, Hong Zhang, Zhibo Chen, Fangwei Zhong, Zhizheng Zhang, and He
  Wang.
\newblock {TrackVLA++}: Unleashing reasoning and memory capabilities in {VLA}
  models for embodied visual tracking.
\newblock In \emph{Proceedings of the IEEE International Conference on Robotics
  and Automation}, 2026{\natexlab{b}}.

\bibitem[Liu et~al.(2024)Liu, Zeng, Ren, Li, Zhang, Yang, Jiang, Li, Yang, Su,
  Zhu, and Zhang]{liu2024groundingdino}
Shilong Liu, Zhaoyang Zeng, Tianhe Ren, Feng Li, Hao Zhang, Jie Yang, Qing
  Jiang, Chunyuan Li, Jianwei Yang, Hang Su, Jun Zhu, and Lei Zhang.
\newblock Grounding {DINO}: Marrying {DINO} with grounded pre-training for
  open-set object detection.
\newblock In \emph{Proceedings of the European Conference on Computer Vision},
  2024.

\bibitem[Maalouf et~al.(2024)Maalouf, Jadhav, Jatavallabhula, Chahine, Vogt,
  Wood, Torralba, and Rus]{maalouf2024follow}
Alaa Maalouf, Ninad Jadhav, Krishna~Murthy Jatavallabhula, Makram Chahine,
  Daniel~M. Vogt, Robert~J. Wood, Antonio Torralba, and Daniela Rus.
\newblock Follow anything: Open-set detection, tracking, and following in
  real-time.
\newblock \emph{IEEE Robotics and Automation Letters}, 9\penalty0 (4):\penalty0
  3283--3290, 2024.

\bibitem[Miao et~al.(2026)Miao, Wei, Ge, Sun, Gao, Zhu, Wang, Tang, Xiao, Tang,
  and Li]{sage3d}
Bingchen Miao, Rong Wei, Zhiqi Ge, Xiaoquan Sun, Shiqi Gao, Jingzhe Zhu, Renhan
  Wang, Siliang Tang, Jun Xiao, Rui Tang, and Juncheng Li.
\newblock Towards physically executable {3D} gaussian for embodied navigation.
\newblock In \emph{The Fourteenth International Conference on Learning
  Representations}, 2026.

\bibitem[{NVIDIA}(n.d.)]{nvidia_isaac_sim_requirements}
{NVIDIA}.
\newblock {Isaac Sim} requirements.
\newblock
  \href{https://docs.isaacsim.omniverse.nvidia.com/latest/installation/requirements.html}{Official
  documentation}, n.d.
\newblock Accessed November 23, 2026.

\bibitem[{OpenBMB}(2026)]{minicpmrobot2026}
{OpenBMB}.
\newblock {MiniCPM-Robot}: A smarter and faster on-device ai brain for robots.
\newblock \href{https://github.com/OpenBMB/MiniCPM-Robot}{Official repository},
  2026.
\newblock Open-source model and code repository.

\bibitem[Oquab et~al.(2024)Oquab, Darcet, Moutakanni, Vo, Szafraniec, Khalidov,
  Fernandez, Haziza, Massa, El-Nouby, Assran, Ballas, Galuba, Howes, Huang, Li,
  Misra, Rabbat, Sharma, Synnaeve, Xu, J{\'e}gou, Mairal, Labatut, Joulin, and
  Bojanowski]{oquab2024dinov2}
Maxime Oquab, Timoth{\'e}e Darcet, Th{\'e}o Moutakanni, Huy~V. Vo, Marc
  Szafraniec, Vasil Khalidov, Pierre Fernandez, Daniel Haziza, Francisco Massa,
  Alaaeldin El-Nouby, Mahmoud Assran, Nicolas Ballas, Wojciech Galuba, Russell
  Howes, Po-Yao Huang, Shang-Wen Li, Ishan Misra, Michael Rabbat, Vasu Sharma,
  Gabriel Synnaeve, Hu Xu, Herv{\'e} J{\'e}gou, Julien Mairal, Patrick Labatut,
  Armand Joulin, and Piotr Bojanowski.
\newblock {DINOv2}: Learning robust visual features without supervision.
\newblock \emph{Transactions on Machine Learning Research}, 2024.

\bibitem[Qiu and Yuille(2016)]{qiu2016unrealcv}
Weichao Qiu and Alan Yuille.
\newblock {UnrealCV}: Connecting computer vision to unreal engine.
\newblock In \emph{European conference on computer vision}, pages 909--916.
  Springer, 2016.

\bibitem[{Qwen Team}(2026{\natexlab{a}})]{qwen38_27b}
{Qwen Team}.
\newblock {Qwen3.8-27B}.
\newblock \href{https://huggingface.co/Qwen/Qwen3.8-27B}{Official model card},
  2026{\natexlab{a}}.

\bibitem[{Qwen Team}(2026{\natexlab{b}})]{qwenrobotnav2026}
{Qwen Team}.
\newblock {Qwen-RobotNav} technical report: A scalable navigation model
  designed for an agentic navigation system.
\newblock \href{https://arxiv.org/abs/2606.18112v3}{arXiv:2606.18112},
  2026{\natexlab{b}}.

\bibitem[Ross et~al.(2011)Ross, Gordon, and Bagnell]{ross2011dagger}
Stephane Ross, Geoffrey Gordon, and Drew Bagnell.
\newblock A reduction of imitation learning and structured prediction to
  no-regret online learning.
\newblock In \emph{Proceedings of the Fourteenth International Conference on
  Artificial Intelligence and Statistics}, pages 627--635, 2011.

\bibitem[Rubner et~al.(2000)Rubner, Tomasi, and Guibas]{rubner2000emd}
Yossi Rubner, Carlo Tomasi, and Leonidas~J. Guibas.
\newblock The earth mover's distance as a metric for image retrieval.
\newblock \emph{International Journal of Computer Vision}, 40\penalty0
  (2):\penalty0 99--121, 2000.

\bibitem[Seiitkazy et~al.(2026)Seiitkazy, Makhanov, Karmanov, Utemuratov,
  Kossakov, Tumanova, and Kim]{seiitkazy2026camera}
Nurgul Seiitkazy, Mukhtar Makhanov, Darkhan Karmanov, Jenis Utemuratov, Marlen
  Kossakov, Mansiya Tumanova, and Alexander Kim.
\newblock Camera-only closed-loop human following by a {Boston Dynamics Spot}:
  A system-level experimental study of detection, tracking, monocular ranging,
  and velocity control.
\newblock \emph{Robotics}, 15\penalty0 (11):\penalty0 214, 2026.

\bibitem[Sun et~al.(2026)Sun, Zhou, Gao, Zhang, Hu, Wen, Ye, and
  Tan]{oavat2026}
Haowei Sun, Kai Zhou, Hao Gao, Shiteng Zhang, Jinwu Hu, Xutao Wen, Qixiang Ye,
  and Mingkui Tan.
\newblock Instance-level visual active tracking with occlusion-aware planning.
\newblock In \emph{Proceedings of the IEEE/CVF Conference on Computer Vision
  and Pattern Recognition}, pages 42494--42504, 2026.

\bibitem[{Ultralytics}(n.d.)]{ultralyticsyolov5}
{Ultralytics}.
\newblock {YOLOv5}.
\newblock \href{https://github.com/ultralytics/yolov5}{Official repository},
  n.d.

\bibitem[Wang et~al.(2025)Wang, Zhang, Li, Liu, Li, Wu, Zhong, Yu, Zhang, and
  Wang]{trackvla2025}
Shaoan Wang, Jiazhao Zhang, Minghan Li, Jiahang Liu, Anqi Li, Kui Wu, Fangwei
  Zhong, Junzhi Yu, Zhizheng Zhang, and He Wang.
\newblock {TrackVLA}: Embodied visual tracking in the wild.
\newblock In \emph{Proceedings of the 9th Conference on Robot Learning}, pages
  4139--4164, 2025.

\bibitem[Wang et~al.(2026)Wang, Luo, Huang, Xu, Wang, Wang, Ma, Yang, Mei, Wei,
  Hu, Liu, Chen, Shao, Lin, Li, Pan, Liu, Ma, and Fan]{lightnav2026}
Shaoan Wang, Aocheng Luo, Fei Huang, Jingyi Xu, Xiaoyang Wang, Yueyu Wang,
  Qianli Ma, Fan Yang, Ran Mei, Jia Wei, Jiangpeng Hu, Xuhao Liu, Hongming
  Chen, Yuanbin Shao, Yiyang Lin, Ziliang Li, Liang Pan, Xinhang Liu, Yuntao
  Ma, and Tingxiang Fan.
\newblock {LightNav-0}: Eliciting {VLM} spatial intelligence for generalist
  embodied navigation.
\newblock \href{https://arxiv.org/abs/2608.30935}{arXiv:2608.30935}, 2026.

\bibitem[Wen et~al.(2026)Wen, Zeng, Peng, Situ, Ye, and Zhang]{wen2026vpevt}
Yanci Wen, Tianle Zeng, Jianwei Peng, Weixi Situ, Hanjing Ye, and Hong Zhang.
\newblock {VP-EVT}: Active visibility preservation for embodied visual
  tracking, 2026.

\bibitem[Wu et~al.(2026)Wu, Chen, Han, Liu, Wang, Wang, Li, Liu, and
  Zhong]{adatracker2026}
Kui Wu, Hao Chen, Jinzhu Han, Haijun Liu, Churan Wang, Yizhou Wang, Zhoujun Li,
  Si Liu, and Fangwei Zhong.
\newblock {AdaTracker}: Learning adaptive in-context policy for
  cross-embodiment active visual tracking.
\newblock \emph{IEEE Robotics and Automation Letters}, 11\penalty0
  (7):\penalty0 8308--8314, 2026.

\bibitem[Wu et~al.(2023)Wu, Escontrela, Hafner, Abbeel, and
  Goldberg]{wu2023daydreamer}
Philipp Wu, Alejandro Escontrela, Danijar Hafner, Pieter Abbeel, and Ken
  Goldberg.
\newblock {DayDreamer}: World models for physical robot learning.
\newblock In \emph{Proceedings of The 6th Conference on Robot Learning}, pages
  2226--2240, 2023.

\bibitem[Wu et~al.(2025)Wu, Wang, Yang, Liu, Zeng, Ye, and Li]{wu2025ortrack}
You Wu, Xucheng Wang, Xiangyang Yang, Mengyuan Liu, Dan Zeng, Hengzhou Ye, and
  Shuiwang Li.
\newblock Learning occlusion-robust vision transformers for real-time {UAV}
  tracking.
\newblock In \emph{Proceedings of the IEEE/CVF Conference on Computer Vision
  and Pattern Recognition}, pages 17103--17113, 2025.

\bibitem[Yao et~al.(2025)Yao, Gao, and Xu]{yao2025navmorph}
Xuan Yao, Junyu Gao, and Changsheng Xu.
\newblock {NavMorph}: A self-evolving world model for vision-and-language
  navigation in continuous environments.
\newblock In \emph{Proceedings of the IEEE/CVF International Conference on
  Computer Vision}, pages 5536--5546, 2025.

\bibitem[Ye et~al.(2022)Ye, Chang, Ma, Shan, and Chen]{ye2022ostrack}
Botao Ye, Hong Chang, Bingpeng Ma, Shiguang Shan, and Xilin Chen.
\newblock Joint feature learning and relation modeling for tracking: A
  one-stream framework.
\newblock In \emph{European Conference on Computer Vision}, 2022.

\bibitem[Ye et~al.(2025)Ye, Situ, Peng, Zhan, Xia, Cai, and
  Zhang]{ye2025followbench}
Hanjing Ye, Weixi Situ, Jianwei Peng, Yu Zhan, Bingyi Xia, Kuanqi Cai, and Hong
  Zhang.
\newblock {Follow-Bench}: A unified motion planning benchmark for
  socially-aware robot person following, 2025.

\bibitem[Ye et~al.(2026{\natexlab{a}})Ye, Zeng, Zhang, Wang, Zhang, Situ, Zhou,
  Ling, and Zhang]{refertrack2026}
Hanjing Ye, Tianle Zeng, Jiazhao Zhang, Shaoan Wang, Zibo Zhang, Weisi Situ,
  Yuchen Zhou, Yonggen Ling, and Hong Zhang.
\newblock {ReferTrack}: Referring then tracking for embodied visual tracking.
\newblock \href{https://arxiv.org/abs/2607.20061}{arXiv:2607.20061},
  2026{\natexlab{a}}.

\bibitem[Ye et~al.(2026{\natexlab{b}})Ye, Zhan, Situ, Chen, Yu, Zhao, Cai,
  Ajoudani, and Zhang]{ye2025tpt}
Hanjing Ye, Yu Zhan, Weixi Situ, Guangcheng Chen, Jingwen Yu, Ziqi Zhao, Kuanqi
  Cai, Arash Ajoudani, and Hong Zhang.
\newblock {TPT-Bench}: A large-scale, long-term and robot-egocentric dataset
  for benchmarking target person tracking.
\newblock \emph{The International Journal of Robotics Research},
  2026{\natexlab{b}}.

\bibitem[Zeng et~al.(2024)Zeng, Zhang, Ehsani, Hendrix, Salvador, Herrasti,
  Girshick, Kembhavi, and Weihs]{zeng2024poliformer}
Kuo-Hao Zeng, Zichen Zhang, Kiana Ehsani, Rose Hendrix, Jordi Salvador, Alvaro
  Herrasti, Ross Girshick, Aniruddha Kembhavi, and Luca Weihs.
\newblock {PoliFormer}: Scaling on-policy {RL} with transformers results in
  masterful navigators.
\newblock In \emph{Proceedings of the 8th Conference on Robot Learning}, pages
  408--432, 2024.

\bibitem[Zhang et~al.(2025)Zhang, Wang, Wang, Li, Liu, Wei, Wang, Zhang, and
  Wang]{zhang2024uninavid}
Jiazhao Zhang, Kunyu Wang, Shaoan Wang, Minghan Li, Haoran Liu, Songlin Wei,
  Zhongyuan Wang, Zhizheng Zhang, and He Wang.
\newblock {Uni-NaVid}: A video-based vision-language-action model for unifying
  embodied navigation tasks.
\newblock In \emph{Proceedings of Robotics: Science and Systems}, 2025.

\bibitem[Zhang et~al.(2026{\natexlab{a}})Zhang, Li, Qi, Li, Liu, Wang,
  et~al.]{zhang2025navfom}
Jiazhao Zhang, Anqi Li, Yunpeng Qi, Minghan Li, Jiahang Liu, Shaoan Wang,
  et~al.
\newblock Embodied navigation foundation model.
\newblock In \emph{International Conference on Learning Representations},
  2026{\natexlab{a}}.

\bibitem[Zhang et~al.(2026{\natexlab{b}})Zhang, Gong, Hao, Fu, Zhang, Zhou, Ye,
  Liang, Liang, and Ding]{zhang2026futurenav}
Lingfeng Zhang, Zeying Gong, Xiaoshuai Hao, Haoxiang Fu, Qiang Zhang, Mingliang
  Zhou, Hangjun Ye, Xiaojun Liang, Junwei Liang, and Wenbo Ding.
\newblock {FutureNav}: Unified world-action modeling for vision-and-language
  navigation, 2026{\natexlab{b}}.

\bibitem[Zhang et~al.(2019)Zhang, Song, Rong, and Li]{zhang2018coarse}
Wei Zhang, Ke Song, Xuewen Rong, and Yibin Li.
\newblock Coarse-to-fine {UAV} target tracking with deep reinforcement
  learning.
\newblock \emph{IEEE Transactions on Automation Science and Engineering},
  16\penalty0 (4):\penalty0 1522--1530, 2019.

\bibitem[Zhang et~al.(2023)Zhang, Ghimire, Javed, Dias, and
  Werghi]{zhang2023robot}
Xiaoxiong Zhang, Adarsh Ghimire, Sajid Javed, Jorge Dias, and Naoufel Werghi.
\newblock Robot-person tracking in uniform appearance scenarios: A new dataset
  and challenges.
\newblock \emph{IEEE Transactions on Human-Machine Systems}, 53\penalty0
  (3):\penalty0 549--559, 2023.

\bibitem[Zhong et~al.(2019)Zhong, Sun, Luo, Yan, and Wang]{zhong2019advat}
Fangwei Zhong, Peng Sun, Wenhan Luo, Tingyun Yan, and Yizhou Wang.
\newblock {AD-VAT}: An asymmetric dueling mechanism for learning visual active
  tracking.
\newblock In \emph{International Conference on Learning Representations}, 2019.

\bibitem[Zhong et~al.(2021)Zhong, Sun, Luo, Yan, and Wang]{zhong2021advatplus}
Fangwei Zhong, Peng Sun, Wenhan Luo, Tingyun Yan, and Yizhou Wang.
\newblock {AD-VAT+}: An asymmetric dueling mechanism for learning and
  understanding visual active tracking.
\newblock \emph{IEEE Transactions on Pattern Analysis and Machine
  Intelligence}, 43\penalty0 (5):\penalty0 1467--1482, 2021.

\bibitem[Zhong et~al.(2023)Zhong, Bi, Zhang, Zhang, and Wang]{zhong2023rspt}
Fangwei Zhong, Xiao Bi, Yudi Zhang, Wei Zhang, and Yizhou Wang.
\newblock {RSPT}: Reconstruct surroundings and predict trajectory for
  generalizable active object tracking.
\newblock In \emph{Proceedings of the AAAI Conference on Artificial
  Intelligence}, pages 3705--3714, 2023.

\bibitem[Zhong et~al.(2024)Zhong, Wu, Ci, Wang, and Chen]{zhong2024offlineevt}
Fangwei Zhong, Kui Wu, Hai Ci, Churan Wang, and Hao Chen.
\newblock Empowering embodied visual tracking with visual foundation models and
  offline {RL}.
\newblock In \emph{Proceedings of the European Conference on Computer Vision},
  2024.

\end{thebibliography}
}

\clearpage
\appendix
\setcounter{table}{0}
\setcounter{figure}{0}
\setcounter{equation}{0}
\renewcommand{\thesection}{\Alph{section}}
\renewcommand{\thetable}{\Alph{table}}
\renewcommand{\thefigure}{\Alph{figure}}
\renewcommand{\theequation}{S\arabic{equation}}
\renewcommand{\theHtable}{S\Alph{table}}
\renewcommand{\theHfigure}{S\Alph{figure}}
\renewcommand{\theHequation}{S\arabic{equation}}

\maketitlesupplementary

\section{Implementation and Hyperparameters}
\label{sec:supp_implementation}

\subsection{Implementation Details}
\label{sec:implementation_details_v4}

\subsubsection{Inference Details}
At inference step $t$, the policy uses the causal RGB window
$I_{t-h:t}$ with $h=31$ and the first-frame target template
$\operatorname{Crop}(I_0,B_0)$, where $B_0$ is the initial target
bounding box.
A shared, frozen DINOv2 ViT-B/14 encoder with register
tokens~\cite{oquab2024dinov2,darcet2024registers} extracts features at
layer indices 2, 5, 8, and 11. We concatenate these features and apply
the DINO-Foresight PCA projection~\cite{karypidis2025dinoforesight}
to obtain 1,152-dimensional representations. The World Expert uses
a frozen feature predictor $\mathcal{P}$ and current- and future-state
readouts. The predictor forecasts $\widetilde{F}_{t+1}$ from the
$N_p=4$ most recent feature maps. Learned adapters project the visual features to
384 dimensions, yielding $C_t$ and $\hat{C}_{t+1}$ for the current and
forecast maps. A four-layer causal Transformer encodes per-frame
global tokens with relative-time encodings and learned temporal
position embeddings. We add the resulting representations
to the spatial tokens to form the visual memory $Y_0$.

The first of the $M=3$ Track Perceiver Blocks receives $N_q=64$
learned tracking queries projected from 2,560 to 384 dimensions.
The blocks fuse the queries with $Y_0$ and the target template tokens
$E^{\mathrm{tar}}$, producing query outputs $q_j$ at each stage.
Using the pooled final queries $\bar{q}$, template features, and
temporal context, the current-state readout $R_{\mathrm{cur}}$ estimates
$\hat{s}_t$ from $C_t$, and the future-state readout $R_{\mathrm{fut}}$
predicts $\hat{s}_{t+1}$ from $\hat{C}_{t+1}$. The future-state readout
also receives the executed command history $a_{t-h:t-1}$. Both readouts
output target visibility, an image-plane point distribution
over a $48 \times 48$ grid with an additional class for an invisible
target, and a robot-frame 3D target point with position uncertainty.

We project the action context $g_t^{\mathrm{ctx}}$ and the per-stage
queries $q_j$ to form the 2,560-dimensional conditioning tokens
$\mathcal{G}_t$. The Action Expert has 16 layers, a hidden width of
1,024, and 32 attention heads. Starting from Gaussian noise, it
integrates the conditional flow field~\cite{lipman2023flow} with four
Euler steps to generate $N_a=8$ normalized planar pose increments
$(\Delta x, \Delta y, \Delta\theta)$ in the current robot frame.
Each increment spans $0.1\,\mathrm{s}$, giving a total horizon of
$0.8\,\mathrm{s}$. We denormalize and cumulatively sum the increments
to obtain the local trajectory $a_t$. A deterministic $\mathrm{SE}(2)$
controller uses the first denormalized increment and the control
interval $\Delta t$ to compute the velocity command. We normalize
the command $(v_f, v_l, \omega_{\mathrm{yaw}})$ and clip each
component to $[-1,1]$.

\subsubsection{Training Details}
Training proceeds in two stages. The visual encoder and feature
predictor remain frozen throughout. In Stage~1, we train the feature
adapters, causal temporal encoder, Track Perceiver Blocks, and
target-state readouts with spatial supervision using data from
TPT-Bench~\cite{ye2025tpt}, \dataset{}, and
EVT-Bench~\cite{trackvla2025}. Specifically, this supervision trains
both current- and future-state readouts to predict the three spatial
outputs described above: target visibility, the $48 \times 48$
image-plane point distribution (including the invisible class), and
the robot-frame 3D target point with position uncertainty. In Stage~2,
we focus on training the Action Expert with expert supervision
and online DAgger~\cite{ross2011dagger}.

At each control step, the Action Expert predicts $N_a=8$ planar pose
increments. We normalize the forward displacement, lateral
displacement, and yaw increment.
Let $X_t^\star$ denote the sequence of normalized ground-truth
increments and $M_t$ its validity mask. We sample noise
$\epsilon \sim \mathcal{N}(0,I)$ and flow time
$\tau = 0.999(1-U^{2/3})$, where $U \sim \mathcal{U}(0,1)$.
At control step $t$ and flow time $\tau$, the interpolated trajectory is
$X_{t,\tau} = (1-\tau)\epsilon + \tau X_t^\star$. Its target flow
velocity is $V_t^\star = X_t^\star-\epsilon$. We minimize
\begin{equation}
\mathcal{L}_{\mathrm{flow}} =
\frac{\sum_{j,d} M_{t,jd}
\bigl[f_\theta(X_{t,\tau},\tau,\mathcal{G}_t)_{jd}
      -V_{t,jd}^\star\bigr]^2}
{\sum_{j,d} M_{t,jd}}.
\label{eq:implementation_flow}
\end{equation}

\subsubsection{Compute Infrastructure}
We use eight NVIDIA H100 GPUs for model training and four NVIDIA
RTX 4090 GPUs for \dataset{} rollouts and closed-loop evaluation
in Isaac Sim, which requires GPUs with RT
cores~\cite{nvidia_isaac_sim_requirements}.

\subsection{Hyperparameter Analysis}

\subsubsection{Hyperparameter Settings}

\label{sec:default_settings}

Table~\ref{tab:trackwam_default_settings} summarizes the default
configuration of \method{} using the notation from the main paper.
The RGB history contains $h$ past frames and the current frame.
The frozen feature predictor $\mathcal{P}$ uses the $N_p$ most recent
feature maps to forecast the next-frame feature map. For action
generation, $N_a$ denotes the number of planar pose increments in
the predicted trajectory, and $K_{\mathrm{Euler}}$ denotes the number
of Euler integration steps.

\begin{table}[t]
    \centering
    \caption{\textbf{Default hyperparameter settings of \method{}.}}
    \label{tab:trackwam_default_settings}
    \footnotesize
    \setlength{\tabcolsep}{3pt}
    \begin{tabularx}{\columnwidth}{@{}l
        >{\raggedright\arraybackslash}X
        >{\raggedright\arraybackslash}p{0.30\columnwidth}@{}}
        \toprule
        Symbol & Quantity & Default setting \\
        \midrule
        $h$ & Number of past RGB frames & 31 \\
        $N_p$ & Feature predictor context length & 4 \\
        $M$ & Number of Track Perceiver Blocks & 3 \\
        $N_q$ & Number of learned tracking queries & 64 \\
        $p_{t:t+1}$ & Image-plane point distributions
            & $48 \times 48$ \\
        $\mathcal{G}_t$ & Action conditioning token width
            & 2,560 \\
        $N_a$ & Number of trajectory increments & 8 \\
        $K_{\mathrm{Euler}}$ & Number of Euler integration steps & 4 \\
        \bottomrule
    \end{tabularx}
\end{table}

\subsubsection{Sensitivity Analysis}

\label{sec:inference_sensitivity}

We vary the RGB window length $h+1$ and the number of Euler steps
$K_{\mathrm{Euler}}$ independently at inference time. We use the
same trained model weights and EVT-Bench STT episodes for all runs.
To compare history lengths, we use 8, 16, or 32 RGB frames and fix
$K_{\mathrm{Euler}}=4$. We use the same frame sampling interval,
target initialization, and four-frame input length for the feature
predictor. To compare step counts, we use 2, 4, or 8 integration steps
with a 32-frame RGB window and the same trajectory horizon and
controller. All settings use batch size 1 on an RTX~4090, matching
the hardware model and batch size in Table~\ref{tab:forktrack_results} of the main paper.

\noindent\textbf{RGB history length.}
In Table~\ref{tab:inference_sensitivity}, increasing the window from
8 to 32 frames raises SR from 82.7\% to 92.2\% and TR from
83.9\% to 89.7\%. Mean latency ranges from 65.42 to
68.20\,ms across the three window lengths.

\noindent\textbf{Euler integration steps.}
Using two steps reduces mean latency from 68.20 to 53.06\,ms,
with SR and TR decreasing by 0.8 and 0.3 percentage points,
respectively. Using eight steps increases latency to 93.40\,ms.
Four steps give the highest SR and TR among the tested step counts.
Our default uses a 32-frame window and four Euler steps.

\begin{table}[t]
    \centering
    \caption{\textbf{Sensitivity to RGB history length and Euler
    integration steps on EVT-Bench STT.} The shaded row indicates
    the default. SR, TR, and CR are percentages. Latency is the mean
    per control step within this batch-1 sweep.}
    \label{tab:inference_sensitivity}
    \footnotesize
    \setlength{\tabcolsep}{3pt}
    \begin{tabularx}{\columnwidth}{@{}
        >{\centering\arraybackslash}X
        >{\centering\arraybackslash}X
        cccc@{}}
        \toprule
        RGB window length ($h+1$) & Euler steps & SR$\uparrow$ & TR$\uparrow$
            & CR$\downarrow$ & Lat. (ms)$\downarrow$ \\
        \midrule
        \rowcolor{cvprblue!12}
        32 & 4 & \textbf{92.2} & \textbf{89.7} & 0.7 & 68.20 \\
         8 & 4 & 82.7 & 83.9 & \textbf{0.5} & 65.42 \\
        16 & 4 & 90.7 & 88.4 & 0.8 & 67.76 \\
        32 & 2 & 91.4 & 89.4 & 1.1 & \textbf{53.06} \\
        32 & 8 & 91.4 & 89.4 & 1.3 & 93.40 \\
        \bottomrule
    \end{tabularx}
\end{table}

\section{Evaluation Protocols and Baseline Details}
\label{sec:supp_protocols}

\subsection{Robot Setup}
EVT-Bench uses Spot in Habitat~\cite{trackvla2025}, while
\dataset{} uses Go2 in Isaac Sim. We match the nominal RGB camera
configuration (Table~\ref{tab:stt_protocol_comparison}), normalized
command interface, and thresholds for target visibility. Both
benchmarks measure following distance between the robot and target
root positions in 3D. The benchmarks differ in scene assets,
collision geometry, and early termination rules. Go2's smaller
footprint can provide more clearance along the same path.

\begin{table}[t]
    \centering
    \caption{\textbf{Robot dimensions and RGB camera configuration.}
    Go2 has a smaller nominal body. Both benchmarks match the camera
    height, forward offset, and horizontal field of view.}
    \label{tab:stt_protocol_comparison}
    \footnotesize
    \setlength{\tabcolsep}{4pt}
    \begin{tabularx}{\columnwidth}{@{}
        >{\raggedright\arraybackslash}Xcc@{}}
        \toprule
        \textbf{Setting} & \textbf{EVT-Bench} & \textbf{\dataset{}} \\
        \midrule
        Robot & Spot & Go2 \\
        Body ($L{\times}W{\times}H$, m)
            & $1.10 \times 0.50 \times 0.84$
            & $0.70 \times 0.31 \times 0.40$ \\
        Camera height (m) & 0.72 & 0.72 \\
        Camera forward offset (m) & 0.24 & 0.24 \\
        Horizontal field of view & $90^\circ$ & $90^\circ$ \\
        \bottomrule
    \end{tabularx}
\end{table}

\subsection{Target Initialization}
\label{sec:supp_target_init}

In the reported EVT-Bench runs of \method{}, the first-frame box
$B_0$ is the axis-aligned bounding rectangle of the target's
ground-truth panoptic mask from the simulator. We also derive per-frame training
box labels from ground-truth masks. At inference, \method{} uses
$B_0$ to crop the fixed target template from $I_0$ and updates its
tracking state from incoming RGB frames and executed commands.

A text-conditioned detector such as Grounding DINO~\cite{liu2024groundingdino}
can instead supply $B_0$ from the first image and the instruction.
Both inputs feed the same tracking policy through the initial
template. Section~\ref{sec:supp_first_frame_grounding} evaluates
this first-frame localization step on EVT-Bench.

\subsection{Baseline Methods}

On EVT-Bench, we use published results for EVT and PoliFormer and
evaluate OA-VAT and ORTrack+PID under the official protocol. On
\dataset{}, we evaluate all four methods alongside the open-source
MLLM-based baselines under a unified protocol. In our evaluations,
we use the released weights as provided. For MLLM-based baselines, we
use their released inference pipelines with the benchmark's
observation and action interfaces. We use the released OmTrackVLA policy,
the predecessor evaluated as a baseline in the VLX-Go
report~\cite{lee2026vlxgorobustembodiedtracking}. The EVT and PoliFormer
implementations below describe our \dataset{} runs.

\noindent\textbf{Offline EVT.}
We run the pretrained policy~\cite{zhong2024offlineevt} with its
original DEVA, Grounding DINO, and SAM pipeline to generate target
masks from simulator RGB frames. We map policy outputs to forward
and angular velocities using calibrated fixed gains and apply a
bounded recovery rotation when the target mask is lost.

\noindent\textbf{PoliFormer.}
For PoliFormer~\cite{zeng2024poliformer}, we extract DINOv2 features
online from the simulator RGB frames, detect pedestrians with
Grounding DINO, and associate boxes across frames using appearance
similarity. We map discrete navigation actions to continuous
robot velocities.

\noindent\textbf{ORTrack+PID.}
We initialize a frozen ORTrack tracker~\cite{wu2025ortrack} with
the first-frame target box and feed its subsequent predictions to
OA-VAT's visual PID controller.

\noindent\textbf{OA-VAT.}
We initialize the frozen OA-VAT tracking and planning
pipeline~\cite{oavat2026} with the first-frame target box.
It processes RGB observations thereafter. We map its control
outputs to simulator velocities using calibrated forward and angular
gains.

\section{Additional Simulation Experiments}
\label{sec:supp_results}

\subsection{Sensitivity to UWB Measurement Errors}
\label{sec:uwb_sensitivity}

We extend the main paper's optional UWB ablation on EVT-Bench STT.
The UWB-augmented model encodes target range and bearing as spatial
tokens and appends them to the input sequence before self-attention.
The vision-only model is the reference in Table~\ref{tab:uwb_sensitivity}.
All three UWB settings share trained weights and evaluation episodes.
We add zero-mean Gaussian noise to simulated range and bearing in the
two perturbed settings, independently across components and time
steps. The table lists the noise scales. We simulate zero sensor delay.

Relative to unperturbed UWB, the lower noise level changes each
metric by at most 0.4 percentage points. At the higher noise
level, SR drops by 8.9 percentage points, TR drops by 6.4 points,
and CR rises by 5.9 points. Compared with vision only, unperturbed
UWB yields slightly higher SR, lower TR, and higher CR.

\begin{table}[t]
    \centering
    \caption{\textbf{Tracking performance under UWB measurement
    errors on EVT-Bench STT.} $\sigma_r$ and $\sigma_\theta$ denote range
    and bearing noise standard deviations. SR, TR, and CR are percentages.}
    \label{tab:uwb_sensitivity}
    \footnotesize
    \setlength{\tabcolsep}{3pt}
    \begin{tabularx}{\columnwidth}{@{}
        >{\raggedright\arraybackslash}Xrrr@{}}
        \toprule
        Input setting & SR$\uparrow$ & TR$\uparrow$ & CR$\downarrow$ \\
        \midrule
        Vision only & 92.2 & 89.7 & 0.7 \\
        Unperturbed UWB & 92.5 & 88.6 & 2.4 \\
        UWB ($\sigma_r=0.10\,\mathrm{m}$, $\sigma_\theta=5^\circ$) & 92.1 & 88.2 & 2.8 \\
        UWB ($\sigma_r=0.50\,\mathrm{m}$, $\sigma_\theta=15^\circ$) & 83.6 & 82.2 & 8.3 \\
        \bottomrule
    \end{tabularx}
\end{table}

\subsection{Additional Results on \dataset{}}

\begin{table}[t]
    \centering
    \caption{\textbf{Distracted Tracking (DT) and Branched Future Tracking
    (BFT) on \dataset{}.}
    SR, TR, and CR are percentages.
    \textbf{Bold} marks the best value in each column.
    $\dagger$ indicates use of Grounding DINO~\cite{liu2024groundingdino}.
    MiniCPM denotes MiniCPM-RobotTrack.}
    \label{tab:supp_forktrack_detailed}
    \footnotesize
    \setlength{\tabcolsep}{2pt}
    \begin{tabularx}{\columnwidth}{@{}l*{6}{>{\centering\arraybackslash}X}@{}}
        \toprule
        \multirow{2}{*}{Method}
        & \multicolumn{3}{c}{DT}
        & \multicolumn{3}{c}{BFT} \\
        \cmidrule(lr){2-4}\cmidrule(lr){5-7}
        & SR$\uparrow$ & TR$\uparrow$ & CR$\downarrow$
        & SR$\uparrow$ & TR$\uparrow$ & CR$\downarrow$ \\
        \midrule
        \rowcolor{black!8}[0pt][0pt]
        \multicolumn{7}{@{}l@{}}{\textit{RL-based policies}} \\
        PoliFormer$^{\dagger}$~\cite{zeng2024poliformer}
            & 2.0 & 52.4 & 29.0 & 1.0 & 58.7 & 43.7 \\
        EVT~\cite{zhong2024offlineevt}
            & 2.0 & 62.8 & 28.3 & 0.7 & 74.7 & 71.3 \\
        \midrule
        \rowcolor{black!8}[0pt][0pt]
        \multicolumn{7}{@{}l@{}}{\textit{Modular pipelines}} \\
        ORTrack+PID~\cite{wu2025ortrack,astrom2006pid}
            & 10.7 & 66.4 & 3.3 & 38.7 & 75.9 & \textbf{4.0} \\
        OA-VAT~\cite{oavat2026}
            & 8.5 & 64.1 & \textbf{2.7} & 36.7 & 77.2 & 4.3 \\
        \midrule
        \rowcolor{black!8}[0pt][0pt]
        \multicolumn{7}{@{}l@{}}{\textit{MLLM-based methods}} \\
        Uni-NaVid~\cite{zhang2024uninavid}
            & 5.0 & 70.0 & 27.0 & 38.0 & 87.0 & 30.7 \\
        OmTrackVLA~\cite{lee2026vlxgorobustembodiedtracking}
            & 13.3 & 68.3 & 5.3 & 55.7 & 81.7 & 5.0 \\
        MiniCPM~\cite{minicpmrobot2026}
            & 9.3 & 66.1 & 8.7 & 33.0 & 77.4 & 19.7 \\
        LightNav-0~\cite{lightnav2026}
            & 16.7 & 69.2 & 8.7 & 38.0 & 79.5 & 16.7 \\
        \midrule
        \rowcolor{cvprblue!12}
        \textbf{\method{}}
            & \textbf{33.7} & \textbf{77.2} & 5.3
            & \textbf{56.7} & \textbf{92.5} & 6.3 \\
        \bottomrule
    \end{tabularx}
\end{table}

Table~\ref{tab:supp_forktrack_detailed} extends the main paper's
STT results to closed-loop DT and BFT. \method{} achieves the highest
SR and TR on both tasks.

\begin{samepage}
\noindent\textbf{DT.}
\method{} achieves 33.7\% SR and 77.2\% TR, with an SR gain of
17.0 percentage points over \mbox{LightNav-0}. These gains show more reliable
target following amid similar-looking pedestrians, consistent with the
main paper's appearance-similarity analysis of explicit perceptual
supervision.

\noindent\textbf{BFT.}
\method{} reaches 56.7\% SR and 92.5\% TR, improving over the best
baselines for each metric by 1.0 and 5.5 percentage points,
respectively. The higher TR indicates more sustained following as
the target takes different routes from a shared history, consistent
with the main paper's paired-branch analysis of the World Expert.

\end{samepage}

\subsubsection{Distractor Appearance Levels}

We quantify target--distractor color dissimilarity using the
1-Wasserstein distance between their CIELAB color distributions
(Lab-W1). The nominal ranges used to select the five distractor
levels are 0--15 for L1, 15--30 for L2, 30--40 for L3, 40--50 for
L4, and $>$ 50 for L5. We vary clothing colors on the same base
avatar for L1--L4 and use a different avatar for L5.
Table~\ref{tab:supp_lab_w1} reports the measured distances for the
two scenes in the main paper.

\begin{table}[t]
    \centering
    \caption{\textbf{Measured target--distractor Lab-W1 distances.}}
    \label{tab:supp_lab_w1}
    \small
    \setlength{\tabcolsep}{2.5pt}
    \begin{tabular}{@{}lrrrrr@{}}
        \toprule
        Scene & L1 & L2 & L3 & L4 & L5 \\
        \midrule
        0139 & 13.220 & 26.674 & 37.770 & 48.588 & 61.928 \\
        0792 &  9.861 & 24.584 & 38.545 & 47.197 & 56.312 \\
        \bottomrule
    \end{tabular}
\end{table}

\subsubsection{Representative BFT Scenes}

\begin{figure}[t]
    \centering
    \includegraphics[width=\linewidth,pagebox=cropbox]
        {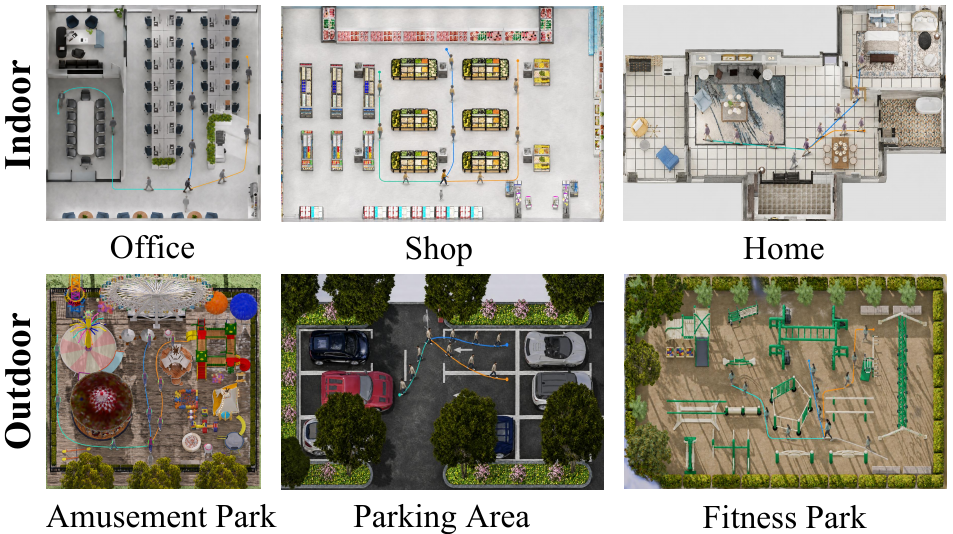}
    \caption{\textbf{Representative indoor and outdoor BFT episodes
    in \dataset{}.} Top-down views show example pedestrian routes
    across six scene types.}
    \label{fig:dataset_example}
\end{figure}

Figure~\ref{fig:dataset_example} shows indoor and outdoor examples
from the BFT split of \dataset{}. For each episode group, we save
the simulator state at the branch point and resume from that state
with different target destinations. The colored routes show the
alternative paths after the shared observation--action history.
We evaluate each continuation in closed loop to measure whether
the policy can follow the target along different routes from the
same starting context.

\subsection{Inference Efficiency}

In Table~\ref{tab:forktrack_results} of the main paper, \method{} takes 68.2\,ms per step
under batch-1 profiling on RTX~4090, 27.7\% less than the fastest
MLLM-based baseline in that comparison (MiniCPM-RobotTrack, 94.3\,ms).
Table~\ref{tab:efficiency_full} provides timing references from
published sources and uses our simulation profiling where no
published timing is available. The offline timings for NavFoM and
TrackVLA++ come from a person recognition benchmark on static
images~\cite{trackvlapp2025}. The real-world entries cover model
inference or the full control loop, as indicated in the table.
For example, the LightNav-0 entry
reports the authors' full control loop on a LimX humanoid. The main
paper reports LightNav-0 model inference in simulation (Table~\ref{tab:forktrack_results})
and on a Unitree Go2 (Table~\ref{tab:real_robot_results}).

\begin{table}[!t]
\centering
\caption{\textbf{Inference and deployment timing.}
Rates are expressed as equivalent intervals.
\emph{This work} denotes batch-1 inference per step in \dataset{}.
Sim.\ and Real denote simulation and real-world deployment,
respectively. Offline denotes person recognition.
$^{*}$ marks end-to-end or closed-loop timing.}
\label{tab:efficiency_full}
\footnotesize
\setlength{\tabcolsep}{1.0pt}
\begin{tabularx}{\columnwidth}{@{}lr@{\hspace{4.5pt}}lc>{\raggedright\arraybackslash}X@{}}
\toprule
Method & Time (ms) & Hardware & Setting & Source \\
\midrule
\rowcolor{black!8}[0pt][0pt]
\multicolumn{5}{@{}l@{}}{\textit{RL-based policies}} \\
PoliFormer & 173.7 & RTX 4090 & Sim. & This work \\
EVT & 66 & RTX 3090 & Sim. & EVT~\cite{zhong2024offlineevt} \\
\midrule
\rowcolor{black!8}[0pt][0pt]
\multicolumn{5}{@{}l@{}}{\textit{Modular pipelines}} \\
ORTrack+PID & 6.8 & RTX 4090 & Sim. & This work \\
OA-VAT & 28.6$^{*}$ & RTX 3090 & Real & OA-VAT~\cite{oavat2026} \\
\midrule
\rowcolor{black!8}[0pt][0pt]
\multicolumn{5}{@{}l@{}}{\textit{MLLM-based methods}} \\
Uni-NaVid & 200.0 & A100 & Real & Uni-NaVid~\cite{zhang2024uninavid} \\
NavFoM & 196.1 & RTX 4090 & Offline & TrackVLA++~\cite{trackvlapp2025} \\
TrackVLA & 100.0 & RTX 4090 & Real & TrackVLA~\cite{trackvla2025} \\
TrackVLA++ & 208.3 & RTX 4090 & Offline & TrackVLA++~\cite{trackvlapp2025} \\
OmTrackVLA & 98.0 & RTX 4090 & Sim. & This work \\
MiniCPM & $\approx 180^{*}$ & Orin NX & Real & \mbox{MiniCPM-Robot~\cite{minicpmrobot2026}} \\
Qwen-RobotNav & 204$^{*}$ & AGX Thor & Real & \mbox{Qwen-RobotNav~\cite{qwenrobotnav2026}} \\
ABot-N0 & 500.0 & Orin NX & Real & ABot-N0~\cite{abotn02026} \\
LightNav-0 & 208.3--222.2$^{*}$ & AGX Thor & Real & LightNav-0~\cite{lightnavthor2026} \\
\midrule
\rowcolor{cvprblue!12}
\textbf{\method{}} & 68.2 & RTX 4090 & Sim. & This work \\
\bottomrule
\end{tabularx}
\end{table}

\subsection{One-Time Language Grounding}
\label{sec:supp_first_frame_grounding}

MLLM-based methods, such as TrackVLA~\cite{trackvla2025}, process
language instructions and RGB history at every control step, whereas
\method{} uses a single initial bounding box to specify the target
throughout the episode. While our main closed-loop rollouts use
simulator-derived initial boxes (Section~\ref{sec:supp_target_init}),
we examine here two practical questions: whether language
descriptions alone can unambiguously identify the target on the
first frame, and whether language grounding can be decoupled into a
one-time startup step.

Table~\ref{tab:first_frame_grounding} evaluates 1,405 first frames
per task. Success requires selecting the designated person and
reaching the Intersection over Union (IoU)~\cite{everingham2010pascal}
threshold with the offline instance box. We compare lightweight
open-vocabulary detectors (Grounding DINO~\cite{liu2024groundingdino}),
direct bounding-box prediction by MLLMs (Qwen3-VL-4B~\cite{bai2025qwen3vl}
and Qwen3.8-27B~\cite{qwen38_27b}), and a two-stage setup that ranks
YOLOv5s~\cite{ultralyticsyolov5} person proposals using either
detection confidence or text-conditioned MLLM selection.

\begin{table}[!t]
    \centering
    \caption{\textbf{First-frame target grounding on EVT-Bench.}
    All values are percentages of the 1,405 first frames per task.
    SR@$\tau_{\mathrm{IoU}}$ requires the correct person and
    IoU $\geq\tau_{\mathrm{IoU}}$. Bold marks the best SR@0.5 per task.
    \emph{Wrong}: wrong-person selection; \emph{No box}: no valid
    box. Single-person STT omits two-stage ranking.}
    \label{tab:first_frame_grounding}
    \footnotesize
    \setlength{\tabcolsep}{2pt}
    \begin{tabularx}{\columnwidth}{@{}>{\raggedright\arraybackslash}Xrrrr@{}}
        \toprule
        Method & SR@0.5 & SR@0.75 & Wrong & No box \\
        \midrule
        \rowcolor{black!8}[0pt][0pt]
        \multicolumn{5}{@{}l@{}}{\textit{STT (Single-Target Tracking)}} \\
        Qwen3.8-27B & 42.92 & 11.32 & 0.00 & 8.33 \\
        Grounding DINO & 89.75 & 88.47 & 0.00 & 0.00 \\
        YOLOv5s & 98.58 & 97.30 & 0.00 & 0.93 \\
        Qwen3-VL-4B & \textbf{99.43} & 95.80 & 0.00 & 0.21 \\
        \midrule
        \rowcolor{black!8}[0pt][0pt]
        \multicolumn{5}{@{}l@{}}{\textit{DT (Distracted Tracking)}} \\
        Qwen3.8-27B & 48.33 & 13.67 & 2.63 & 4.20 \\
        YOLOv5s & 71.25 & 70.25 & 27.54 & 0.57 \\
        Grounding DINO & 78.29 & 76.80 & 8.47 & 0.14 \\
        YOLOv5s + Qwen3-VL-4B & 93.38 & 91.81 & 5.48 & 0.57 \\
        YOLOv5s + Qwen3.8-27B & 94.23 & 92.67 & 4.70 & 0.57 \\
        Qwen3-VL-4B & \textbf{96.37} & 92.31 & 3.06 & 0.28 \\
        \midrule
        \rowcolor{black!8}[0pt][0pt]
        \multicolumn{5}{@{}l@{}}{\textit{AT (Ambiguity Tracking)}} \\
        Qwen3.8-27B & 48.68 & 14.16 & 0.00 & 0.28 \\
        YOLOv5s & 96.87 & 95.80 & 1.85 & 0.85 \\
        YOLOv5s + Qwen3-VL-4B & 97.94 & 96.87 & 0.64 & 0.85 \\
        YOLOv5s + Qwen3.8-27B & 98.65 & 97.58 & 0.07 & 0.85 \\
        Grounding DINO & 99.57 & 99.15 & 0.00 & 0.07 \\
        Qwen3-VL-4B & \textbf{99.64} & 95.02 & 0.00 & 0.14 \\
        \bottomrule
    \end{tabularx}
\end{table}

Three findings emerge from Table~\ref{tab:first_frame_grounding}.
First, text-conditioned grounding remains susceptible to instance confusion
in multi-person scenes: even the strongest direct model,
Qwen3-VL-4B, selects the wrong person in 3.06\% (43 of 1,405) of
DT frames, contributing to a 3.63\% (51 of 1,405) overall first-frame
failure rate on DT (whereas an initial box specifies the target
instance directly). Across YOLOv5s proposals, adding
text-conditioned selection improves SR@0.5 from 71.25\% to
93.38--94.23\% on DT and from 96.87\% to 97.94--98.65\% on AT.
Second, decoupling language grounding to the initial frame is both
effective and lightweight: Grounding DINO achieves 78.29\% on DT and
99.57\% on AT in a single forward pass, while Qwen3-VL-4B reaches
96.37\% on DT and over 99.4\% on STT and AT. Once initialized,
the RGB-only configuration of \method{} relies solely on the
cropped target template, causal RGB history, and executed commands,
avoiding per-step MLLM inference overhead in the closed-loop
control cycle.
Third, larger parameter scale alone does not ensure accurate direct
bounding-box localization: Qwen3.8-27B achieves only 42.92--48.68\%
SR@0.5 (and 11.32--14.16\% SR@0.75) despite low wrong-person rates
(0.00--2.63\%), indicating weak coordinate regression rather than
semantic confusion. Delegating box proposals to YOLOv5s raises its
SR@0.5 to 94.23\% on DT and 98.65\% on AT, surpassing
YOLOv5s + Qwen3-VL-4B and showing that grounding alignment or
two-stage proposal ranking matters more than model size alone.

\begin{table}[!t]
    \centering
    \caption{\textbf{Closed-loop recovery with optional UWB after
    first-frame grounding failures on EVT-Bench.} We evaluate the
    43 episodes (5/8 STT, 35/51 DT, 3/5 AT) that failed
    Qwen3-VL-4B's SR@0.5 yet succeeded with ground-truth boxes.
    Episodes without a box cannot initialize the tracker; \method{}
    (+ UWB) recovers 33/34 initialized cases (97.1\%) and 33/43
    (76.7\%) overall.}
    \label{tab:grounding_recovery}
    \footnotesize
    \setlength{\tabcolsep}{4pt}
    \begin{tabularx}{\columnwidth}{@{}>{\raggedright\arraybackslash}Xcr@{}}
        \toprule
        First-frame grounding failure & + UWB Recovered & Recovery (\%) \\
        \midrule
        No valid box predicted & 0 / 9 & 0.0 \\
        Wrong person selected & 28 / 29 & 96.6 \\
        Valid box (IoU $< 0.5$) & 5 / 5 & 100.0 \\
        \midrule
        \textbf{Total} & \textbf{33 / 43} & \textbf{76.7} \\
        \bottomrule
    \end{tabularx}
    \vspace{-4pt}
\end{table}

\noindent\textbf{Closed-loop evaluation under imperfect initialization.}
We select 43 episodes across the three tasks that count as failures
in Qwen3-VL-4B's first-frame SR@0.5 and succeed in the original
closed-loop evaluation with ground-truth-derived boxes (5/8 STT,
35/51 DT, and 3/5 AT). This motivates our optional UWB interface:
while vision-only tracking is general, it cannot self-correct once
initialized on a distractor crop, whereas lightweight UWB cues
complement vision to provide practical fault tolerance against
imperfect first-frame grounding. We therefore rerun valid-box
episodes under the optional UWB-augmented configuration
(Section~\ref{sec:uwb_sensitivity}). The 9 episodes without a valid
box count as closed-loop failures because the tracker cannot be
initialized. With UWB, the tracker succeeds in 28 of the 29 DT
episodes initialized with a distractor box and in all five other
episodes with a valid box. In total, 33 of 43 episodes (76.7\%)
succeed in this selected subset (Table~\ref{tab:grounding_recovery}),
with language used only for first-frame grounding.

\section{Failure Analysis and Limitations}
\label{sec:failure_analysis}

\begin{figure*}[!t]
    \centering
    \includegraphics[width=\linewidth,pagebox=cropbox]{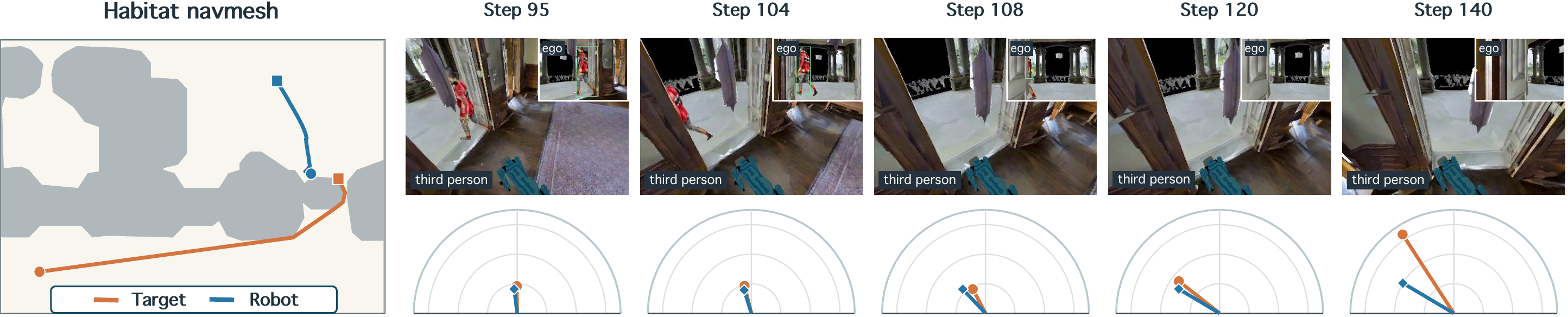}
    \caption{\textbf{Excessive-distance termination at a narrow opening.}
    The Habitat navigation mesh shows the target and robot paths.
    The insets show egocentric views. The polar plots show the growing
    separation as the target continues along its route.}
    \label{fig:failure_case}
    \vspace{-6pt}
\end{figure*}

\begin{figure}[!t]
    \centering
    \includegraphics[width=\columnwidth,pagebox=cropbox]{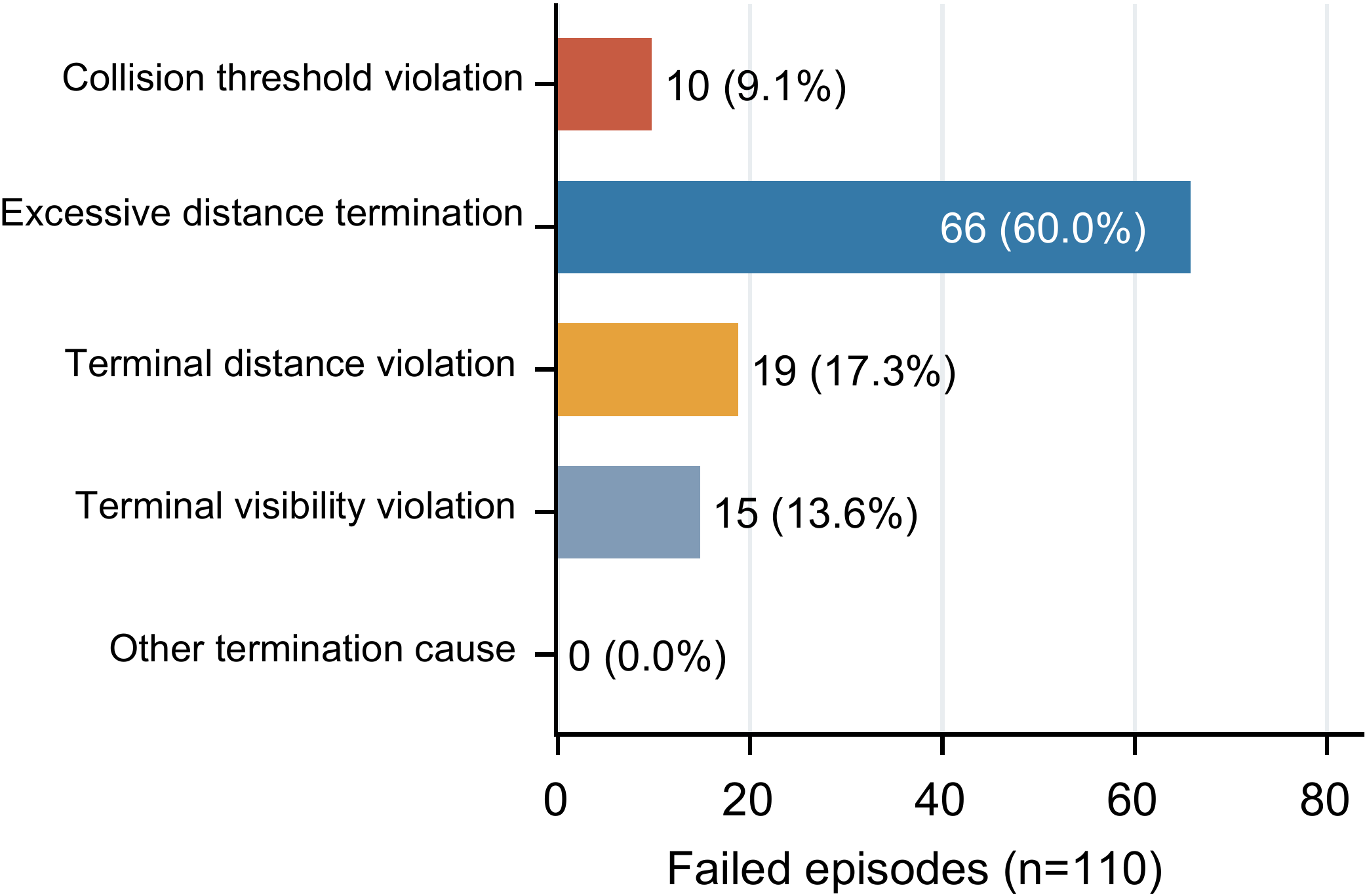}
    \caption{\textbf{Failure outcomes in single-target tracking.}
    Percentages are computed over the 110 failed episodes.}
    \label{fig:fixedset_primary}
    \vspace{-6pt}
\end{figure}

\subsection{Evaluation Outcomes}

Following prior work on robot person following~\cite{seiitkazy2026camera},
we assign each failed episode to the first applicable outcome as follows.

\begin{enumerate}
    \item \textbf{Collision threshold violation.}
    The robot--target distance falls below $0.5\,\mathrm{m}$.
    \item \textbf{Excessive distance termination.}
    The evaluator ends the episode because the target is too far away.
    \item \textbf{Terminal distance violation.}
    The final robot--target distance lies outside $[1,3]\,\mathrm{m}$.
    \item \textbf{Terminal visibility violation.}
    The final distance is valid, but the facing or field-of-view check fails.
    \item \textbf{Other termination cause.}
    The episode ends unsuccessfully for any remaining reason.
\end{enumerate}

We observe 110 failures across 1,405 EVT-Bench single-target tracking episodes.
Excessive distance accounts for 66 failures (60.0\%),
the most common outcome in Figure~\ref{fig:fixedset_primary}.

\subsection{Excessive-Distance Termination}

We find that in 33 of the 66 excessive-distance failures (50.0\%), the robot
moves less than $0.15\,\mathrm{m}$ cumulatively in the horizontal
plane over the final ten steps despite a mean normalized forward
command greater than 0.15. This pattern is consistent with stalled
progress near termination.

Figure~\ref{fig:failure_case} illustrates this failure mode in Habitat.
The target is still visible at step 104 as it passes through a
narrow opening. The robot makes little progress near the opening.
By step 108, the target has left the robot's field of view. It
continues walking until their separation exceeds the evaluation limit.

When a passage appears open in RGB but offers limited clearance,
the vision-only policy can issue forward commands with little
resulting motion. The target can move beyond the following range
before the robot recovers. Explicit motion feedback could help
detect stalled progress and trigger an earlier recovery maneuver.

\section{Additional Real-World Experiments}
\label{sec:supp_robot}

\begin{figure}[!t]
    \centering
    \includegraphics[width=0.68\linewidth,pagebox=cropbox]{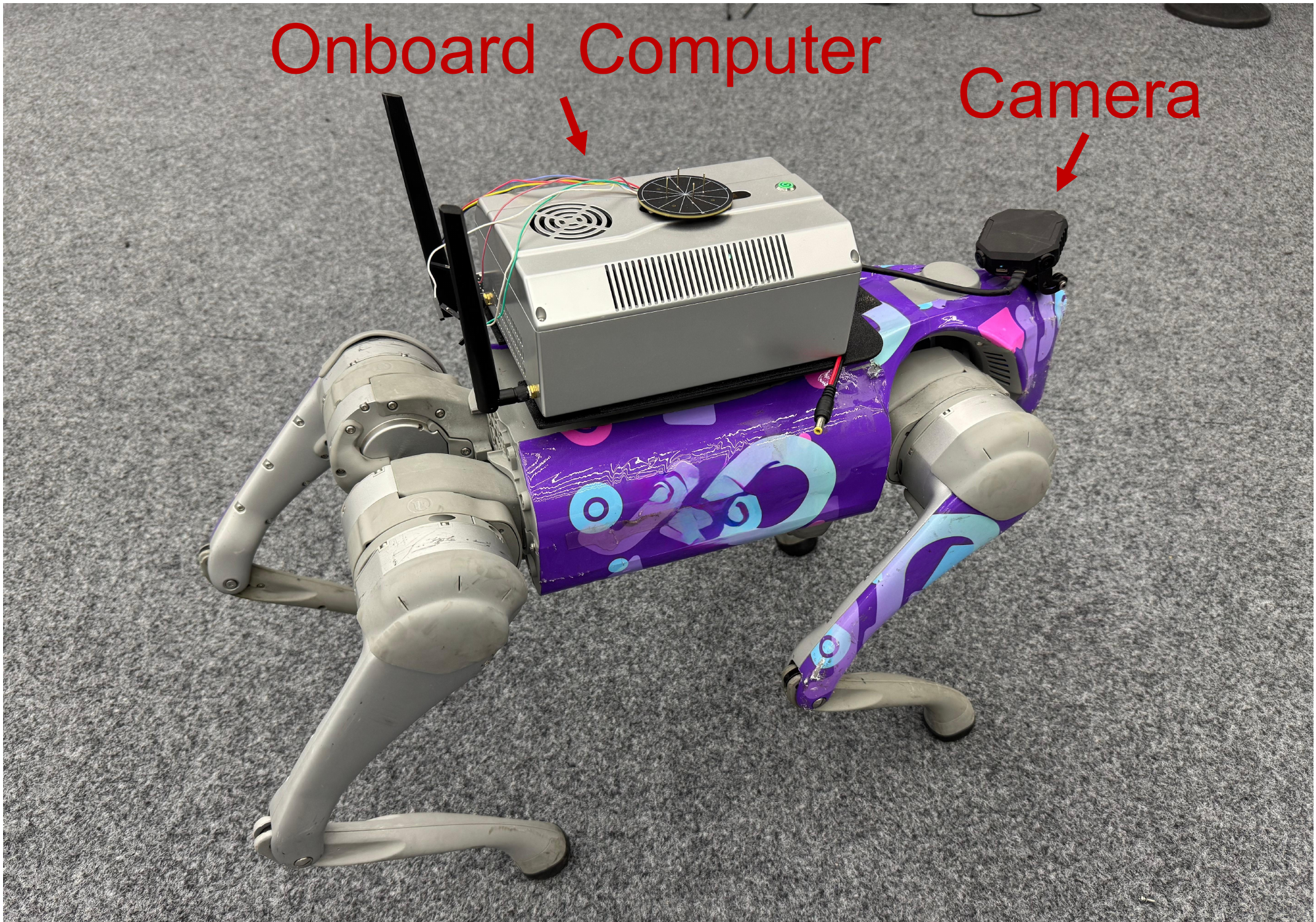}
    \caption{\textbf{Real-world robot platform.}
    The Unitree Go2 carries a forward-facing RGB camera and an NVIDIA
    Jetson AGX Thor onboard computer.}
    \label{fig:real_robot_platform}
    \vspace{-6pt}
\end{figure}

\begin{figure}[!t]
    \centering
    \includegraphics[width=0.90\linewidth,pagebox=cropbox]{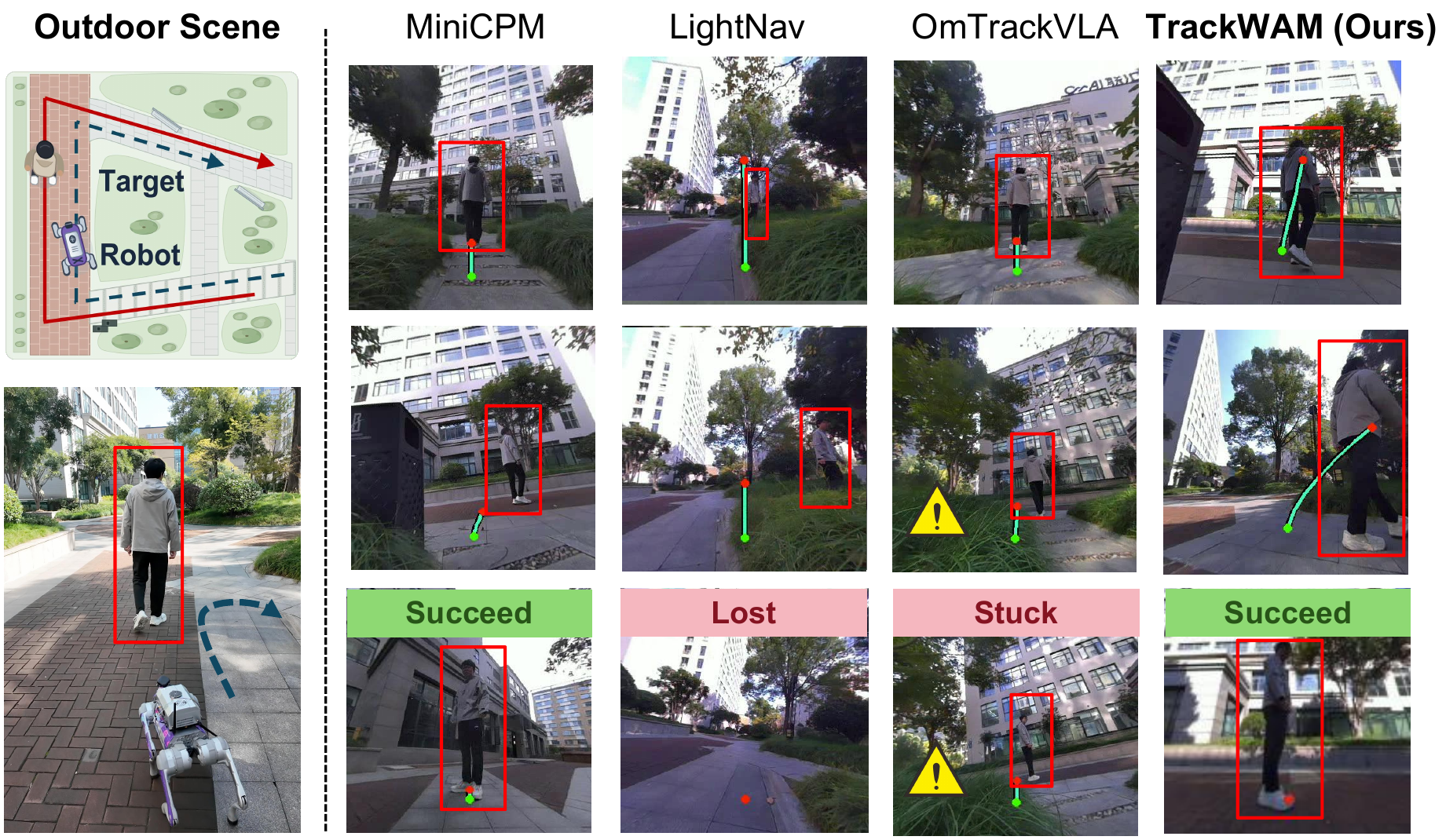}
    \caption{\textbf{Outdoor walkway tracking on Unitree Go2.}}
    \label{fig:real_robot_walkway}
    \vspace{-6pt}
\end{figure}

\begin{figure}[!t]
    \centering
    \includegraphics[width=0.90\linewidth,pagebox=cropbox]{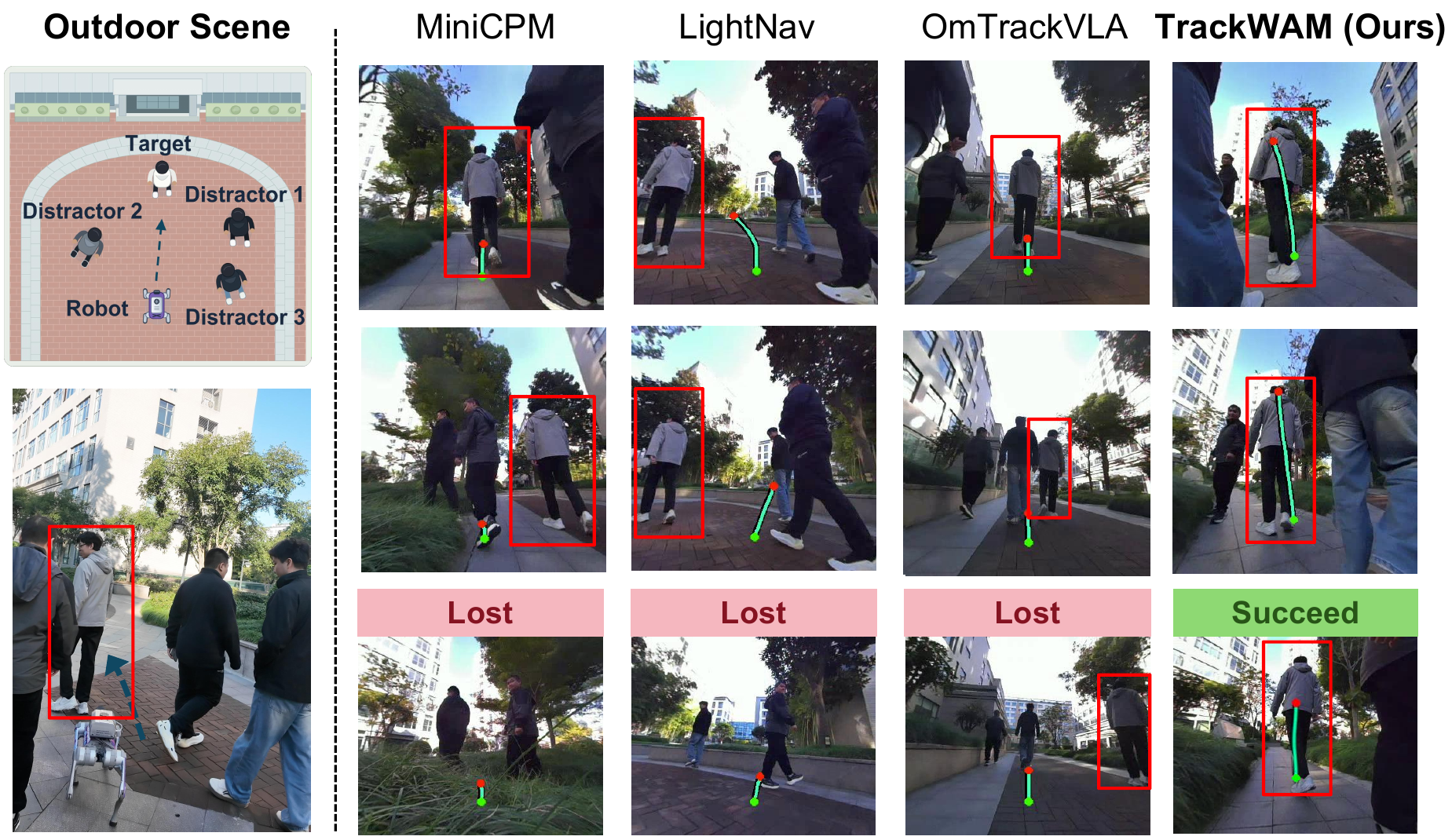}
    \caption{\textbf{Outdoor target following with pedestrian distractors.}}
    \label{fig:real_robot_crowd}
    \vspace{-6pt}
\end{figure}

\subsection{Experimental Setup}

We mount a Seeker Omni-Depth four-camera system above the Unitree
Go2's head, tilted upward by $15^\circ$ (Figure~\ref{fig:real_robot_platform}).
We use only the forward-facing RGB camera, undistorting images
using its intrinsics and cropping them to $384 \times 384$ pixels
with a $90^\circ$ field of view.

All methods run in Docker on Jetson AGX Thor (T5000) in MAXN mode.
LightNav-0~\cite{lightnav2026} uses FP8 quantization, MiniCPM uses
its official TensorRT implementation, and OmTrackVLA and \method{}
use FP16 inference. All real-world trials run vision-only \method{}
(no UWB), using Qwen3-VL-4B~\cite{bai2025qwen3vl} once on frame~0
for $B_0$.

\subsection{Ethical Considerations}
All participants in our real-world experiments volunteered and
provided informed consent. Our institution confirmed that these
internal engineering evaluation trials did not require formal ethics
review.

\subsection{Outdoor Qualitative Results}

The following outdoor sequences complement the indoor trials in
the main paper.

\noindent\textbf{Outdoor walkway.}
The open walkway in Figure~\ref{fig:real_robot_walkway} provides more
space for maneuvering than the narrow indoor route in the main paper.
Both \method{} and MiniCPM complete the illustrated runs. LightNav-0
loses the target, as in the indoor example, while OmTrackVLA takes a
shortcut across the grass and becomes stuck. \method{} maintains
stable following along the walkway, completing all five trials in
Table~\ref{tab:real_robot_results} of the main paper. Its lower inference latency facilitates
timely adjustments to target motion during closed-loop execution.

\noindent\textbf{Pedestrian distractors.}
In Figure~\ref{fig:real_robot_crowd}, the target wears a white top and
black pants, while the three distractors wear a black top with black
pants, a gray top with black pants, and a black top with blue pants.
The quadruped's low viewpoint emphasizes the lower body, making
pedestrians with similar pants hard to distinguish.
OmTrackVLA is confused by all three distractors, whereas MiniCPM and
LightNav-0 are mainly confused by the two in black pants.
\method{} keeps tracking the target throughout the run, aligning with
the main paper's DT analysis on the benefits of explicit spatial
supervision and temporal context under ambiguity and temporary
occlusion.

\end{document}